\documentclass{article}

\usepackage[preprint]{neurips_2026}

\usepackage[utf8]{inputenc}
\usepackage[T1]{fontenc}
\usepackage{hyperref}
\usepackage{url}
\usepackage{booktabs}
\usepackage{amsfonts}
\usepackage{amsmath}
\usepackage{nicefrac}
\usepackage{microtype}
\usepackage[table,dvipsnames]{xcolor}
\usepackage{graphicx}
\usepackage{float}
\usepackage{multirow}
\usepackage{algorithm}
\usepackage{algorithmic}
\usepackage{colortbl}
\usepackage{capt-of}
\definecolor{lightred}{RGB}{255,220,220}
\definecolor{lightgreen}{RGB}{220,255,220}
\definecolor{lightgray}{RGB}{240,240,240}

\newif\ifneuripsversion
\title{Latent Bridge: Feature Delta Prediction for Efficient Dual-System Vision-Language-Action Model Inference}

\author{%
  Yudong Liu\textsuperscript{1} \quad
  Yuan Li\textsuperscript{2} \quad
  Zijia Tang\textsuperscript{1} \quad
  Yuxi Zheng\textsuperscript{1} \quad
  Yueqian Lin\textsuperscript{1} \\
  \bfseries Qinsi Wang\textsuperscript{1} \quad
  Yi Li\textsuperscript{2} \quad
  Shuangjun Liu\textsuperscript{2} \quad
  Shuai Zhang\textsuperscript{2} \quad
  Taotao Jing\textsuperscript{2} \\
  \bfseries Dashan Gao\textsuperscript{2} \quad
  Ning Bi\textsuperscript{2} \quad
  Jingwei Sun\textsuperscript{3} \quad
  Yiran Chen\textsuperscript{1} \quad
  Hai Li\textsuperscript{1} \\[0.5em]
  \mdseries
  \textsuperscript{1}Duke University \quad
  \textsuperscript{2}Qualcomm AI Research \quad
  \textsuperscript{3}University of Florida \\[0.3em]
  \footnotesize
  \textsuperscript{1}\texttt{\{yudong.liu,\,zijia.tang,\,yuxi.zheng,\,yueqian.lin,\,qinsi.wang,\,yiran.chen,\,hai.li\}@duke.edu} \\
  \textsuperscript{2}\texttt{\{yuali,\,yli35,\,shuangju,\,shuazhan,\,tjing,\,dgao,\,nbi\}@qti.qualcomm.com} \\
  \textsuperscript{3}\texttt{sun.jingwei@ufl.edu}
}

\begin{document}

\maketitle

\begin{abstract}
Dual-system Vision-Language-Action (VLA) models achieve state-of-the-art robotic manipulation but are bottlenecked by the VLM backbone, which must execute at every control step while producing temporally redundant features.
We propose \textbf{Latent Bridge}, a lightweight model that predicts VLM output deltas between timesteps, enabling the action head to operate on predicted outputs while the expensive VLM backbone is called only periodically.
We instantiate Latent Bridge on two architecturally distinct VLAs: \textbf{GR00T-N1.6} (feature-space bridge) and \textbf{$\pi_{0.5}$} (KV-cache bridge), demonstrating that the approach generalizes across VLA designs.
Our task-agnostic DAgger training pipeline transfers across benchmarks without modification.
Across four LIBERO suites, 24 RoboCasa kitchen tasks, and the ALOHA sim transfer-cube task, Latent Bridge achieves \textbf{95--100\% performance retention} while reducing VLM calls by \textbf{50--75\%}, yielding \textbf{1.65--1.73$\times$ net per-episode speedup}.
\ifneuripsversion\else
Our code and checkpoints are publicly available\footnote{\url{https://github.com/1999Lyd/Latent-Bridge}}.
\fi
\end{abstract}

% ---- FIGURE 1: Teaser / Overview (top of page 1) ----
\vspace{-3pt}
\begin{figure}[H]
\centering
\includegraphics[width=\textwidth]{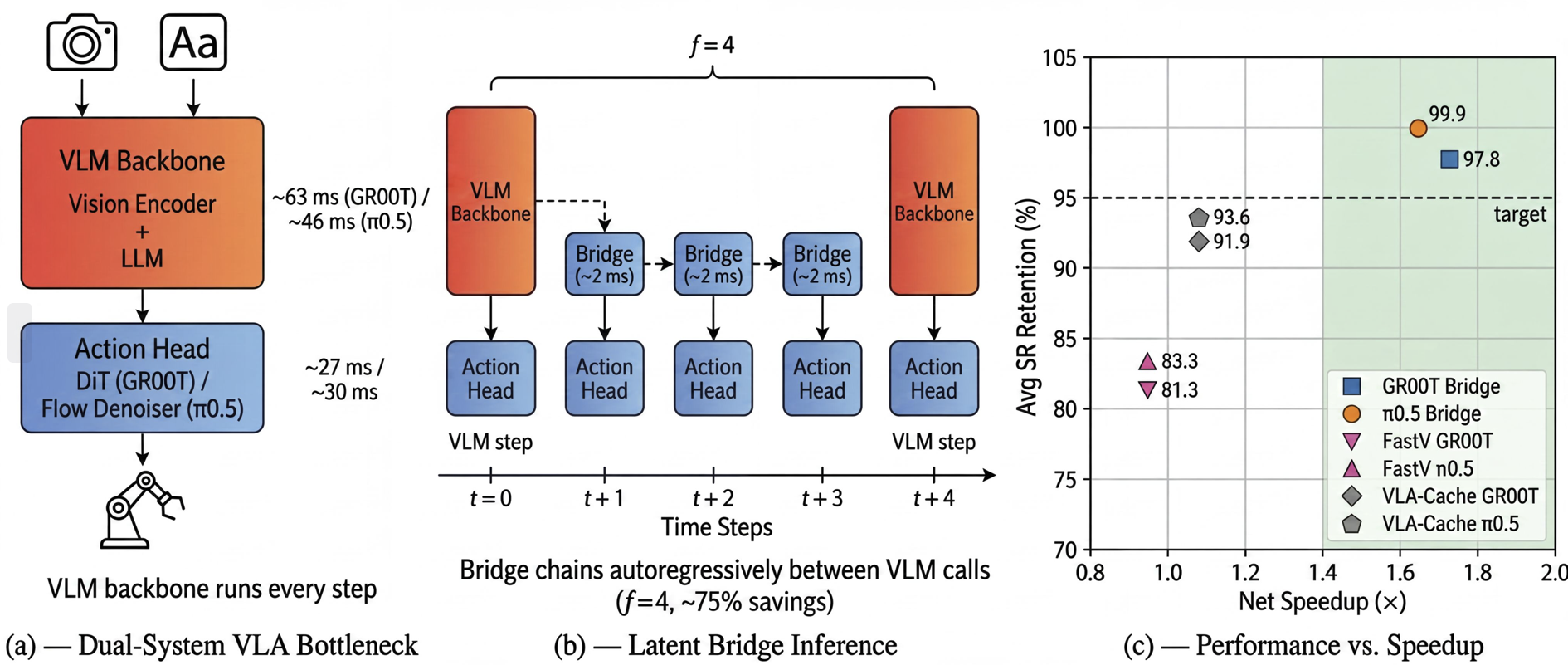}
\caption{Latent Bridge reduces VLM backbone calls by predicting feature deltas between timesteps. The bridge operates at orders-of-magnitude lower latency, enabling 50--75\% VLM savings with 95--100\% task performance retention across two VLA architectures and diverse benchmarks.}
\label{fig:teaser}
\end{figure}
\vspace{-4pt}

%=====================================================================
\section{Introduction}
%=====================================================================

Vision-Language-Action (VLA) models have emerged as a promising paradigm for generalist robotic manipulation, leveraging pretrained vision-language models to ground language instructions in visual observations and produce motor commands~\cite{brohan2023rt2, black2024pi0, bjorck2025groot}.
Among VLA architectures, \emph{dual-system} designs---where a large VLM backbone produces a feature representation that a separate, lightweight action head decodes into actions---have achieved state-of-the-art results on diverse manipulation benchmarks.
Unlike single-system VLAs that autoregressively generate action tokens through the LLM itself (e.g., RT-2~\cite{brohan2023rt2}, OpenVLA~\cite{kim2024openvla}), dual-system VLAs (e.g., GR00T~\cite{bjorck2025groot}, $\pi_{0.5}$~\cite{intelligence2025pi05}, Octo~\cite{team2024octo}) decompose inference into:
\begin{enumerate}
    \item \textbf{Stage 1 (VLM backbone):} A vision encoder + language model processes the observation $o_t$ to produce features $z_t \in \mathbb{R}^{N \times D}$. This is the heavyweight component and dominates inference time.
    \item \textbf{Stage 2 (Action head):} A lightweight policy network decodes $z_t$ into action $a_t$, running significantly faster than Stage~1.
\end{enumerate}

This decomposition creates a fundamental \textbf{asymmetric latency bottleneck}: the VLM backbone must execute at every control step, yet produces outputs that change slowly over time.
For flow-matching action heads (GR00T's DiT~\cite{bjorck2025groot}, $\pi_{0.5}$'s flow expert~\cite{intelligence2025pi05})---both sample the time variable from a $\mathrm{Beta}(1.5, 1.0)$ distribution biased toward the data side---reducing denoising from $\geq$10 steps to a single step incurs negligible performance loss in our experiments, making the VLM backbone the clear bottleneck. For real-time manipulation at 10--50Hz, this severely limits deployment.

\paragraph{Key insight: Feature temporal redundancy.}
We observe that consecutive VLM outputs exhibit high temporal similarity, meaning each expensive backbone call produces features that are largely redundant with the previous timestep.
This redundancy suggests that a much smaller model could \emph{predict} the output delta $\Delta_t = z_{t+1} - z_t$, allowing the action head to operate on predicted features while the VLM backbone is called only every $f$ steps.

\paragraph{Our approach.}
We propose \textbf{Latent Bridge}, a lightweight model that predicts one-step VLM output deltas:
\begin{equation}
    \hat{z}_{t+1} = \hat{z}_t + \mathcal{B}(\hat{z}_t, s_t, q_t, a_{t-1})
\end{equation}
where $\hat{z}_t$ is the most recent representation (the fresh VLM feature on a VLM step, or the bridge's own previous prediction on a chained bridge step), $s_t$ provides observation-dependent context (stable intermediate features or fresh vision embeddings), and $q_t$, $a_{t-1}$ are the robot state and previous action.
Between periodic VLM calls (every $f=2$--$4$ steps), the bridge chains predictions autoregressively, replacing the VLM backbone at orders-of-magnitude lower latency.

A bridge trained on sync data alone---rollouts collected with the VLM running every step, so the bridge always sees clean VLM features as inputs---learns from a clean-input distribution, but at deployment it must consume its own predictions during autoregressive chaining, creating a \emph{distribution shift} that sync-only training does not cover.
We close this gap with a \textbf{task-agnostic DAgger pipeline}: the bridge is rolled out in the simulator while the VLM runs in parallel to provide ground-truth feature targets along the bridge's own trajectory, and the bridge is retrained on this mixed distribution.
This pipeline transfers across all four LIBERO task suites without modification.

Our full pipeline therefore consists of three stages applied task-adaptively: \textbf{Stage~1} is the bridge learning stage (sync collection + R0 supervised training + DAgger R1 refinement) and is the only stage required for simpler benchmarks; \textbf{Stage~2} (LoRA action-head adaptation) and \textbf{Stage~3} (phase-aware VLM scheduling) are added for long-horizon tasks where autoregressive bridge drift compounds over longer rollouts.

Our contributions are:
\begin{enumerate}
    \item \textbf{Latent Bridge}: a lightweight delta predictor that skips $>$50\% of VLM backbone calls on intermediate steps; orthogonal to and composable with existing VLM acceleration methods (pruning, quantization, layer skipping).
    \item We instantiate the bridge on two architecturally distinct VLAs---feature-space for GR00T and KV-cache for $\pi_{0.5}$---and evaluate on LIBERO (4 suites), RoboCasa (24 kitchen tasks), and ALOHA sim (transfer-cube), achieving \textbf{95--100\% SR retention} with \textbf{1.65--1.73$\times$ net speedup}.
    \item A complete three-stage pipeline (bridge learning; LoRA action-head adaptation; phase-aware scheduling), applied task-adaptively---Stage~1 alone recovers near-sync performance on shorter-horizon benchmarks, and all three stages together \textbf{fully recover synchronous VLA performance} on long-horizon tasks---trained on a single GPU in a few hours.
\end{enumerate}

%=====================================================================
\section{Related Work}
%=====================================================================
\vspace{-4pt}
\paragraph{Vision-Language-Action models.}
VLA models combine pretrained vision-language understanding with action prediction for robotic control.
\emph{Single-system} VLAs---RT-1~\cite{brohan2023rt1}, RT-2~\cite{brohan2023rt2}, OpenVLA~\cite{kim2024openvla}, and RT-X~\cite{padalkar2024rtx}---tokenize actions and generate them autoregressively through the LLM, inheriting its sequential decoding bottleneck.
\emph{Dual-system} VLAs decouple perception from action generation: GR00T~\cite{bjorck2025groot} uses a DiT action head, $\pi_{0.5}$~\cite{intelligence2025pi05} uses an interleaved flow-matching expert, Octo~\cite{team2024octo} and HPT~\cite{wang2024hpt} use diffusion decoders, and CogACT~\cite{li2024cogact} uses a continuous action decoder.
Diffusion Policy~\cite{chi2023diffusion} demonstrates that diffusion-based action heads achieve strong multi-modal behavior cloning.
Our work targets the dual-system architecture, exploiting the separable representation between backbone and action head to skip backbone computation on intermediate steps.
\vspace{-10pt}
\paragraph{Efficient VLM/VLA inference.}
General VLM acceleration techniques include KV cache compression~\cite{li2024snapkv, xiao2024streamingllm}, speculative decoding~\cite{leviathan2023fast, cai2024medusa}, visual token pruning~\cite{chen2024fastv, bolya2022token}, and weight quantization~\cite{lin2024awq, frantar2023gptq}. VLA-specific adaptations span token pruning (VLA-Cache~\cite{xu2025vlacache}, SpecPrune-VLA~\cite{wang2025specprune}, LightVLA~\cite{lightvla2025}, VLA-Pruner~\cite{vlapruner2025}, Compressor-VLA~\cite{compressorvla2025}), layer skipping (MoLe-VLA~\cite{molevla2025}), and quantization (SQAP-VLA~\cite{sqapvla2025}, DyQ-VLA~\cite{dyqvla2026}), reporting 1.08--1.62$\times$ \emph{per-step} speedups with various SR trade-offs. None measure net episode wall-clock time, and all still execute the full backbone---making them complementary to, not competitive with, our approach: rather than optimizing the VLM forward pass, we \emph{skip it entirely} on bridge steps, and the bridge composes with any of the above on VLM steps that do execute.
\vspace{-10pt}
\paragraph{World models and latent dynamics.}
World models learn environment dynamics for planning or data augmentation.
Ha \& Schmidhuber~\cite{ha2018world} learn spatial (VAE) and temporal (RNN) representations; the Dreamer line~\cite{hafner2020dreamer, hafner2023dreamerv3} trains policies via backpropagation through latent imagination; IRIS~\cite{micheli2023iris} uses a discrete autoregressive world model with a transformer.
In the visual domain, UniSim~\cite{yang2024unisim} generates realistic experiences via a generative model conditioned on diverse actions, and Genie~\cite{bruce2024genie} produces controllable environments from internet video.
LeCun's JEPA framework~\cite{lecun2022jepa} and V-JEPA~\cite{bardes2024vjepa} advocate prediction in representation space rather than pixel space, which is closest to our bridge's operating principle.
Our bridge similarly predicts future latent states, but in the VLM's hidden representation space rather than a learned abstract state, and serves to accelerate inference rather than enable planning or representation learning.
\vspace{-6pt}

%=====================================================================
\section{Method}
%=====================================================================

\subsection{Problem Formulation}

Consider a dual-system VLA with policy $\pi(a_t \mid o_t) = \mathcal{A}(\mathcal{V}(o_t))$, where $\mathcal{V}$ is the VLM backbone and $\mathcal{A}$ is the action head.
At each control step $t$, the backbone produces features $z_t = \mathcal{V}(o_t) \in \mathbb{R}^{N \times D}$ ($N$: sequence length, $D$: feature dimension), and the action head decodes an action $a_t = \mathcal{A}(z_t, q_t)$ conditioned on proprioceptive state $q_t$.

We seek a lightweight bridge $\mathcal{B}$ that approximates VLM features without running $\mathcal{V}$.
We define the \emph{VLM call period} $f$: the backbone $\mathcal{V}$ executes once every $f$ control steps (i.e., at $t = 0, f, 2f, \ldots$), and the bridge $\mathcal{B}$ fills the remaining $f{-}1$ intermediate steps.
Equivalently, the VLM is invoked at $1/f$ of the control frequency, so the VLM call rate scales as $1/f$.
For example, $f{=}3$ means 1 VLM call followed by 2 bridge steps, reducing VLM compute by $\frac{f-1}{f} = 67\%$.
At intermediate offsets $t' \in \{1, \ldots, f{-}1\}$ from the last VLM call at step $t$, the bridge predicts:
\begin{align}
    \hat{z}_{t+t'} &= \hat{z}_{t+t'-1} + \mathcal{B}\bigl(\hat{z}_{t+t'-1},\; s_t,\; q_{t+t'},\; a_{t+t'-1}\bigr), & t' &= 1, \ldots, f{-}1 \label{eq:bridge} \\
    a_{t+t'} &= \mathcal{A}\bigl(\hat{z}_{t+t'},\; q_{t+t'}\bigr) & &\text{(bridge step action)}  \notag
\end{align}
with the boundary convention $\hat{z}_t := z_t$ (the fresh VLM feature on VLM steps), and where $s_t \in \mathbb{R}^{N \times D}$ denotes stable intermediate-layer VLM features cached from the last VLM call, providing visual context without rerunning the backbone (Section~\ref{sec:arch}).
The bridge therefore chains \emph{autoregressively}: at $t'=1$ the input is the fresh VLM feature $z_t$; at $t' \geq 2$ the input is the bridge's own previous prediction $\hat{z}_{t+t'-1}$.
This autoregressive chaining motivates the DAgger training stage (Section~\ref{sec:training}).

The simplest baseline is \textbf{feature caching}, which sets $\mathcal{B} = \text{Id}$ (i.e., $\hat{z}_{t+1} = z_t$, equivalently $\Delta_t = 0$), reusing stale VLM features on non-VLM steps.
This corresponds to the zero-function bridge and serves as a lower bound; our learned bridge must outperform this trivial predictor.

The period $f$ controls the trade-off between speedup and action quality: larger $f$ reduces VLM calls but accumulates prediction error over more consecutive bridge steps.
Our objective is to find the operating point that maximizes speedup while maintaining $>$95\% task performance retention.

\vspace{-2pt}
\subsection{Latent Bridge Architecture}
\label{sec:arch}

% ---- FIGURE 2: Architecture diagram ----
\vspace{-3pt}
\begin{figure}[t]
\centering
\includegraphics[width=\textwidth]{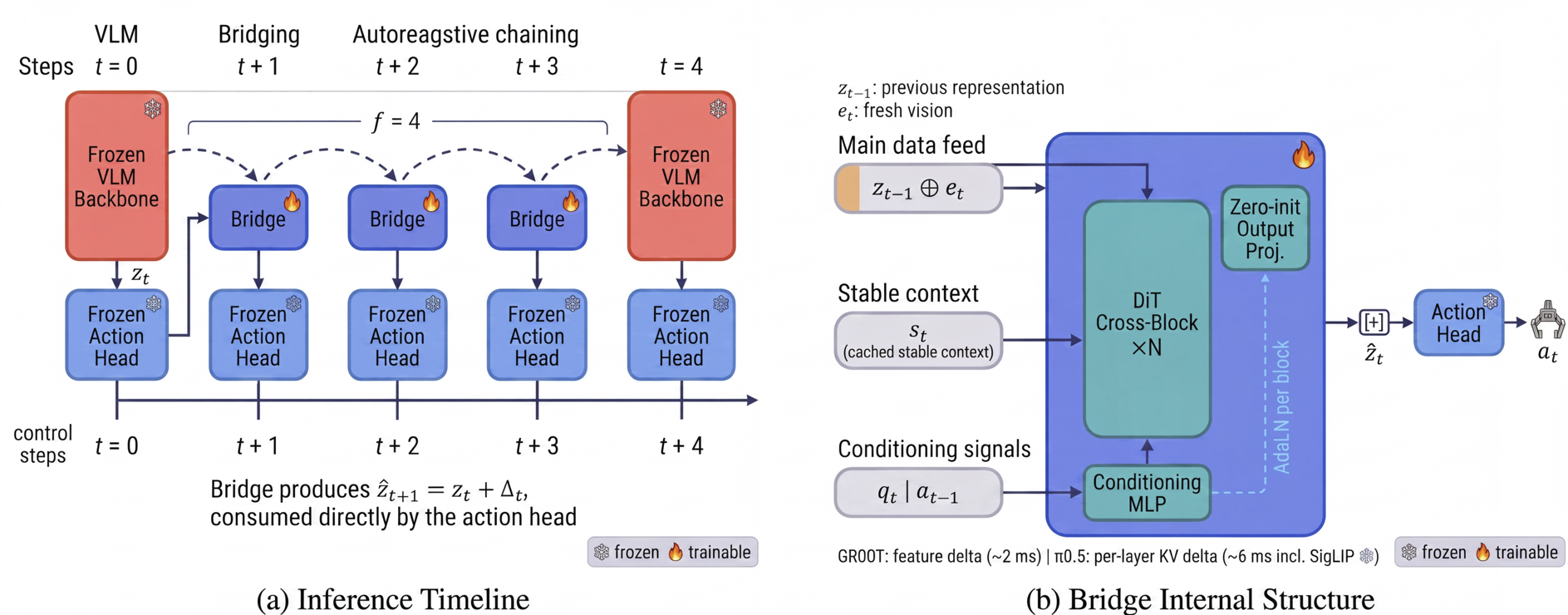}
\caption{Architecture comparison. Both variants use a DiT backbone with AdaLN conditioning. GR00T operates on a single feature vector; $\pi_{0.5}$ operates on per-layer KV pairs. Zero-initialized output ensures the bridge starts at the copy baseline.}
\label{fig:architecture}
\end{figure}
\vspace{-4pt}

\paragraph{Feature-space bridge (GR00T).}
For VLAs with a single feature vector interface between backbone and action head, the bridge predicts one-step feature deltas via residual prediction:
\begin{equation}
    \hat{z}_{t+1} = \hat{z}_t + \Delta_t, \qquad \Delta_t = \mathcal{B}(\hat{z}_t,\; s_t,\; q_t,\; a_{t-1})
    \label{eq:groot_bridge}
\end{equation}
where $\hat{z}_t \in \mathbb{R}^{N_\text{img} \times D}$ is the most recent feature consumed by the action head (fresh from the VLM at a VLM step, or autoregressively chained at a bridge step; boundary $\hat{z}_t := z_t$ as in Eq.~\ref{eq:bridge}), $s_t$ are stable-layer features providing scene context, $q_t$ is the proprioceptive state, and $a_{t-1}$ is the previous action.
The bridge $\mathcal{B}$ is a Transformer with cross-attention (DiT blocks) consisting of:
\begin{itemize}
    \item \textbf{Self-attention} over input features $z_t$ with learned positional embeddings
    \item \textbf{Cross-attention} to stable context $s_t$, providing visual grounding without a vision encoder
    \item \textbf{AdaLN conditioning} on $(q_t, a_{t-1})$, injecting state and action information at every block
    \item \textbf{Zero-initialized output projection}, so the untrained bridge starts at the feature caching baseline ($\Delta_t = 0$)
\end{itemize}
\vspace{-6pt}

\paragraph{Image-only processing.}
Text-token hidden states are nearly invariant between consecutive steps (cosine $>$0.9999) since the instruction is fixed within an episode, so predicting near-zero text deltas would dilute the gradient on image-token deltas. The bridge therefore operates exclusively on image tokens ($N_\text{img} \ll N$) and copies text tokens from the last VLM cache; dropping text from the bridge \emph{input} is safe because $\hat{z}_t$ was already produced by a text-conditioned VLM and implicitly encodes the instruction. Empirically this yields no SR loss while making the bridge sequence-length-agnostic.

\paragraph{Stable vs.\ dynamic layers.}
We decompose the VLM's intermediate hidden representations into \emph{stable} layers (early/middle, cosine $>$0.999 between consecutive steps) and \emph{dynamic} layers (final, cosine $\sim$0.95). The bridge predicts the action-head input $z_t$ and consumes the stable-layer features $s_t$ as cross-attention context, which provide local visual context (spatial layout, object identities) for the current observation. On bridge steps, $s_t$ is reused from the last VLM call while $z_t$ is freshly predicted.

\vspace{-2pt}
\subsection{Training Pipeline}

% ---- FIGURE 3: Training pipeline ----
\vspace{-3pt}
\begin{figure}[t]
\centering
\includegraphics[width=\textwidth]{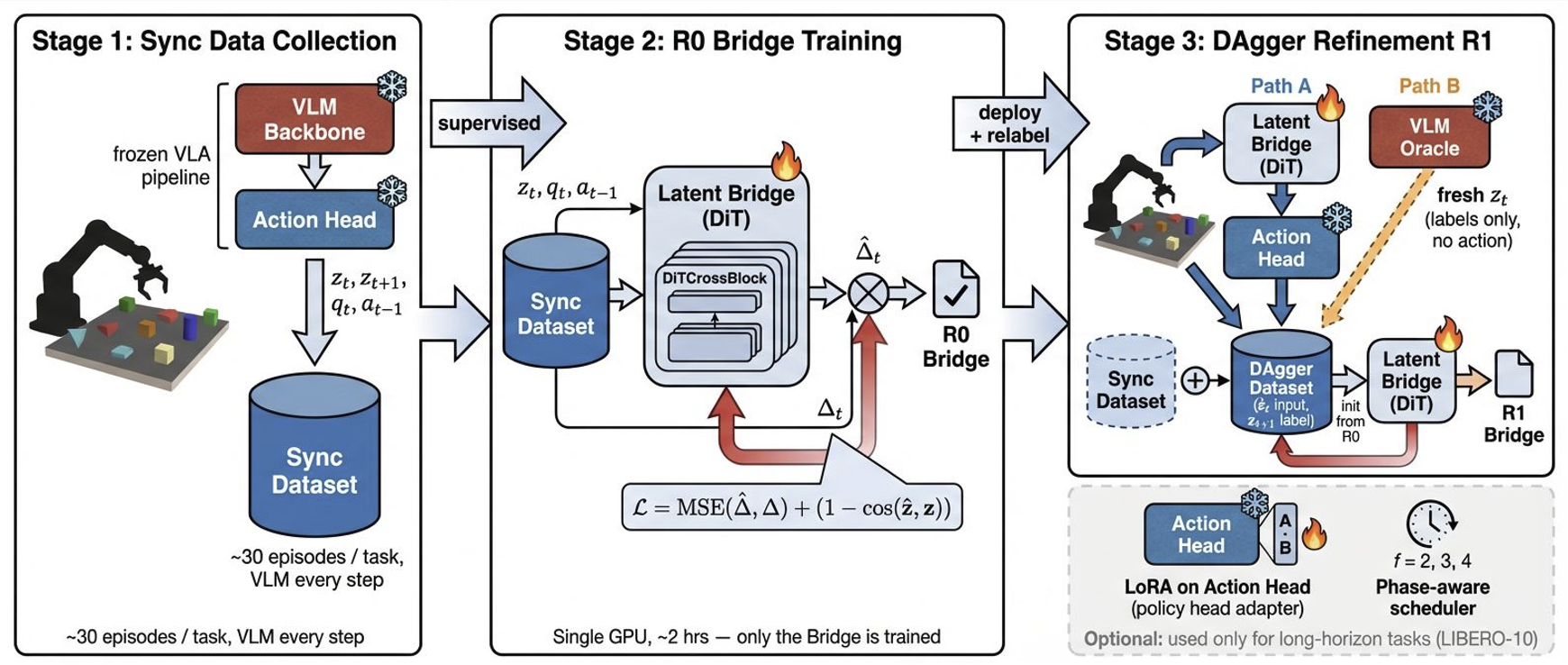}
\caption{Task-agnostic three-stage pipeline. The same pipeline transfers across all LIBERO suites, RoboCasa, and ALOHA sim without modification. DAgger closes the distribution gap between sync training and bridge deployment.}
\label{fig:pipeline}
\end{figure}
\vspace{-4pt}

Our training pipeline has three stages, all task-agnostic:

\paragraph{Stage 1: Sync data collection.}
\label{sec:training}
We roll out the pretrained VLA in sync mode (VLM every step) in the simulator and record tuples $(z_t, z_{t+1}, s_t, q_t, a_{t-1})$ along its closed-loop observation trajectory; bridge training is then fully offline on this fixed dataset.

\paragraph{Stage 2: R0 bridge training.}
We denote the initial bridge trained on sync-only data as the ``round-0'' bridge (\textbf{R0}).
The bridge is trained on sync pairs with a combined MSE and cosine loss, restricted to image tokens:
\begin{equation}
    \mathcal{L} = \| \hat{z}_{t+1} - z_{t+1} \|^2 + \alpha \left(1 - \frac{\hat{z}_{t+1} \cdot z_{t+1}}{\|\hat{z}_{t+1}\| \, \|z_{t+1}\|}\right)
    \label{eq:loss}
\end{equation}
where all terms operate over image tokens only ($\alpha = 1.0$).
Text token deltas are masked from the loss since they are near-zero and would dilute the gradient signal.

\paragraph{Stage 3: DAgger refinement (R1).}
After R0 training, we apply one round of DAgger refinement to produce the ``round-1'' bridge (\textbf{R1}).
The R0 bridge is deployed in simulation with VLM call period $f$, while a VLM oracle runs in parallel to provide ground-truth features at every control step (the robot still follows the bridge policy's actions).
This yields DAgger pairs $(\hat{z}_{t+t'-1},\, z_{t+t'})$ for each bridge step at offset $t' \in \{1,\ldots,f{-}1\}$ from a VLM call at step $t$: the input is exactly what Eq.~\ref{eq:bridge} feeds the bridge---which reduces to the fresh VLM feature $z_t$ at $t'{=}1$ via the boundary convention $\hat{z}_t := z_t$, and is the bridge's own (potentially noisy) prediction at $t' \geq 2$---paired with the oracle's ground-truth feature $z_{t+t'}$ as target.
The bridge is retrained on mixed sync + DAgger data, resuming from R0 weights with a reduced learning rate.
DAgger is essential because at $t' \geq 2$ the bridge's input is its own prediction, not a clean VLM feature; without DAgger, errors compound rapidly at higher $f$.

\vspace{-2pt}
\subsection{Optional Enhancements}
\label{sec:enhancements}

\paragraph{Phase-aware VLM scheduling.}
For complex tasks with diverse motion phases, we dynamically adjust $f$ based on the previous action's translation magnitude $\|a_{t}^{\text{trans}}\|$:
high motion ($\|a_{t}^{\text{trans}}\| > \tau_\text{nav}$) $\rightarrow f{=}2$,
medium ($\tau_\text{manip} < \|a_{t}^{\text{trans}}\| \leq \tau_\text{nav}$) $\rightarrow f{=}3$,
low ($\|a_{t}^{\text{trans}}\| \leq \tau_\text{manip}$) $\rightarrow f{=}4$.
This allocates more VLM compute during high-dynamics phases where bridge predictions are less reliable.

\paragraph{LoRA action head adaptation.}
When bridge features are significantly noisier than clean VLM features, we apply Low-Rank Adaptation (LoRA)~\cite{hu2022lora} to the action head's DiT, training on a 50/50 mix of bridge and sync features with Gaussian noise augmentation.
This increases the action head's tolerance to bridge prediction noise.

\vspace{-2pt}
\subsection{Adaptation to $\pi_{0.5}$: KV-Cache Bridge}

Unlike GR00T's single feature-vector interface, $\pi_{0.5}$~\cite{intelligence2025pi05} uses an interleaved design where the Gemma-2B backbone and Gemma-300M action expert share transformer layers: at each of $L{=}18$ layers, action (suffix) tokens cross-attend to the backbone's (prefix) per-layer KV cache.
We adapt the bridge to predict pre-RoPE key and value deltas for all $L$ layers simultaneously:
\begin{equation}
    \{\Delta K_l, \Delta V_l\}_{l=1}^{L} = \mathcal{B}_\text{KV}(\Delta e_t,\; e_t,\; \widehat{\text{KV}}_{t-1},\; q_t,\; a_{t-1}),
    \label{eq:kv_bridge}
\end{equation}
where $e_t = \text{SigLIP}(o_t)$ is the vision embedding, $\Delta e_t = e_t - e_{t-1}$ encodes what changed visually between steps, and $\widehat{\text{KV}}_{t-1}$ is the most recent per-layer KV cache (fresh from the VLM at a VLM step, or autoregressively chained from the bridge's previous prediction at a bridge step; same boundary convention as Eq.~\ref{eq:bridge}).
$\mathcal{B}_\text{KV}$ uses a shared DiT backbone with 18 lightweight per-layer output heads. SigLIP ($\sim$5ms) provides fresh visual context far more cheaply than the full Gemma prefix ($\sim$46ms); total bridge-step cost is $\sim$6ms with bf16 + \texttt{torch.compile}. Architectural details and RoPE handling are in Appendix~\ref{sec:app_impl}.

\vspace{-2pt}
\subsection{Inference}

At deployment with VLM call period $f$:
\begin{itemize}
    \item \textbf{VLM steps} ($t \bmod f = 0$): Run full backbone $\mathcal{V}$. Cache $z_t$, $s_t$, and KV pairs for bridge use.
    \item \textbf{Bridge steps} ($t \bmod f \neq 0$): $\mathcal{B}$ predicts $\Delta$ from the most recent representation---the fresh VLM feature $z_t$ on the first bridge step, or its own previous output $\hat{z}_{t'-1}$ on subsequent steps (autoregressive chaining). The action head runs on the predicted $\hat{z}_t$. Cost: 2--6ms vs.\ 46--63ms for the replaced VLM backbone.
\end{itemize}

%=====================================================================
\section{Experiments}
%=====================================================================

\subsection{Setup}
\vspace{-4pt}
\paragraph{Base models.}
We evaluate on two dual-system VLAs:
(1)~\textbf{GR00T-N1.6-3B}~\cite{bjorck2025groot}, with an Eagle backbone (SigLIP2 + Qwen3) and DiT action head, fine-tuned per LIBERO suite; and
(2)~\textbf{$\pi_{0.5}$}~\cite{intelligence2025pi05}, with a PaliGemma backbone (SigLIP + Gemma-2B) and Gemma-300M action expert, using the publicly released LIBERO checkpoint.
\vspace{-2pt}
\paragraph{Benchmarks.}
We evaluate on four LIBERO~\cite{liu2024libero} suites:
\textbf{LIBERO-Spatial} (10 spatial reasoning tasks),
\textbf{LIBERO-Object} (10 object manipulation tasks),
\textbf{LIBERO-Goal} (10 goal-conditioned tasks),
and \textbf{LIBERO-10} (10 long-horizon tasks, most challenging).
We additionally evaluate GR00T on \textbf{RoboCasa}~\cite{robocasa2024} (24 kitchen tasks, different robot embodiment, zero-shot) and $\pi_{0.5}$ on \textbf{ALOHA sim}~\cite{zhao2023aloha} (bimanual transfer-cube task) to test cross-platform generalization.
We run 3 random seeds $\times$ 20 episodes per task unless otherwise specified (60 episodes per task; 600 episodes per LIBERO suite of 10 tasks).\footnote{All datasets used in the paper were solely downloaded and evaluated by the university.} All reported success rates are mean across the 3 seeds; the resulting run-to-run standard deviation is below 0.5pp on all reported configurations.
\vspace{-2pt}
\paragraph{Hardware.}
All latency profiling and inference experiments are conducted on a single NVIDIA A100 80GB GPU with bfloat16 precision and \texttt{torch.compile}.
\vspace{-2pt}
\paragraph{Bridge models.}
GR00T bridge: lightweight DiT predicting feature deltas ($\sim$2ms; no vision encoder needed on bridge steps).
$\pi_{0.5}$ bridge: DiT predicting KV-cache deltas ($\sim$6ms; includes SigLIP encoder for fresh embedding input).
Ablation shows that a 19M-parameter bridge achieves identical SR to a 148M bridge; vision input and stable context are both critical for maintaining performance on harder, longer-horizon tasks (Section~\ref{sec:ablation}).
Training: single GPU, a few hours per suite for R0 + DAgger R1.
\vspace{-2pt}
\paragraph{Baselines.}
We compare against FastV~\cite{chen2024fastv} and VLA-Cache~\cite{xu2025vlacache} as token pruning baselines, along with feature caching ($\Delta_t{=}0$).
The majority of VLA pruning methods (VLA-Cache, LightVLA, VLA-Pruner, Compressor-VLA, MoLe-VLA, etc.) are designed and evaluated exclusively on single-system VLAs such as OpenVLA, where the LLM backbone directly generates action tokens.
While SpecPrune-VLA~\cite{wang2025specprune} reports results on $\pi_{0.5}$, it is not open-sourced and involves extensive hyperparameter tuning (action-aware pruning thresholds, layer selection), making faithful reproduction difficult.
In dual-system VLAs, the diffusion or flow-matching action head is sensitive to the dense conditioning provided by VLM hidden states or KV caches, making token pruning adaptation considerably harder than in single-system VLAs where such techniques are already mature.
We select FastV and VLA-Cache as they are open-sourced and reproducible, and we adapt them to dual-system VLAs (GR00T and $\pi_{0.5}$) for fair comparison.

\vspace{-2pt}
\subsection{Main Results}
\vspace{-4pt}
\begin{table}[t]
\caption{\textbf{Comparison of VLA inference methods on LIBERO.} Success rate (\%, mean across 3 random seeds $\times$ 20 episodes per task) across four suites, average per-step latency (averaged over VLM and non-VLM steps within one call period $f$), average episode length on successful episodes, and net wall-clock speedup. Feature Caching ($\Delta_t = 0$) reuses stale VLM outputs at matched $f$ without learned prediction. Full FastV and VLA-Cache adaptation details (layer selection, prune/threshold sweeps) are in Appendix~\ref{sec:app_baselines}.}
\label{tab:main}
\centering
\setlength{\tabcolsep}{3.5pt}
\small
\resizebox{\textwidth}{!}{%
\begin{tabular}{l|cccc|c|ccc}
\toprule
Method & Spatial & Object & Goal & Long & Avg SR$\uparrow$ & Avg Latency (ms)$\downarrow$ & Ep.\ Len$\downarrow$ & Speedup$\uparrow$ \\
\midrule
\multicolumn{9}{l}{\textit{GR00T-N1.6-3B} \quad {\scriptsize VLM Backbone: 63ms \quad Action Head (1-step): 27ms \quad Bridge: 2ms \quad $f{=}2\text{--}4$ (phase-aware)}} \\
\midrule
Sync (every step)        & 96.17 & 99.83 & 97.33 & 93.00 & 96.58 & 90          & 19          & 1.00$\times$ \\
+ FastV                  & 86.17 & 84.00 & 85.00 & 58.83 & 78.50 & 77\,{\scriptsize($-$15\%)} & 23\,{\scriptsize($+$23\%)} & 0.95$\times$ \\
+ VLA-Cache              & 91.17 & 90.00 & 89.00 & 85.17 & 88.84 & 75\,{\scriptsize($-$17\%)} & 21\,{\scriptsize($+$8\%)}  & 1.08$\times$ \\
+ Feature Caching        & 35.67 & 3.00  & 56.00 & 42.33 & 34.25 & 48\,{\scriptsize($-$47\%)} & ---$^\dagger$  & ---$^\dagger$ \\
+ Latent Bridge (ours)   & \textbf{95.83} & \textbf{97.83} & \textbf{95.33} & \textbf{89.17} & \textbf{94.54} & \textbf{49}\,{\scriptsize($-$45\%)} & 20\,{\scriptsize($+$6\%)} & \textbf{1.73$\times$} \\
\midrule
\multicolumn{9}{l}{\textit{$\pi_{0.5}$} \quad {\scriptsize VLM Backbone: 46ms \quad Action Head (1-step): 30ms \quad Bridge: 6ms (SigLIP 5ms + DiT 1ms) \quad $f{=}4$}} \\
\midrule
Sync (every step)        & 98.83 & 98.17 & 97.00 & 93.83 & 96.96 & 76          & 31          & 1.00$\times$ \\
+ FastV                  & 88.17 & 86.00 & 86.83 & 62.17 & 80.79 & 67\,{\scriptsize($-$12\%)} & 37\,{\scriptsize($+$20\%)} & 0.95$\times$ \\
+ VLA-Cache              & 93.17 & 91.00 & 92.17 & 86.83 & 90.79 & 65\,{\scriptsize($-$14\%)} & 33\,{\scriptsize($+$6\%)}  & 1.08$\times$ \\
+ Feature Caching        & 46.67 & 57.33 & 68.83 & 52.67 & 56.38 & 45\,{\scriptsize($-$41\%)} & ---$^\dagger$  & ---$^\dagger$ \\
+ Latent Bridge (ours)   & \textbf{99.00} & \textbf{97.67} & \textbf{97.33} & \textbf{93.67} & \textbf{96.92} & \textbf{46}\,{\scriptsize($-$39\%)} & 33\,{\scriptsize($+$5\%)} & \textbf{1.65$\times$} \\
\bottomrule
\multicolumn{9}{l}{\scriptsize $^\dagger$\,Feature Caching: low SR, most episodes fail to max steps.} \\
\multicolumn{9}{l}{\scriptsize Ep.\ Len = avg control steps per successful episode (GR00T: 1 step = 8 env steps; $\pi_{0.5}$: 1 step = 5 env steps).}
\end{tabular}%
}
\end{table}
\vspace{-4pt}

Latent Bridge achieves \textbf{95--100\% retention} (avg 94.5\% GR00T, 96.9\% $\pi_{0.5}$) while feature caching collapses to 3--56\% SR. Token-pruning baselines still run the full backbone: FastV yields net slowdown (0.95$\times$) and VLA-Cache only 1.08$\times$. These methods report $\sim$1.3--1.5$\times$ on single-system VLAs, but on dual-system VLAs the speedup only affects the backbone fraction (60--70\% of per-step cost), driving the net gain to a low Amdahl ceiling (Appendix~\ref{sec:app_baselines}). Latent Bridge \emph{skips the backbone entirely}---replacing a 46--63\,ms VLM call with a 2--6\,ms bridge---yielding \textbf{1.65--1.73$\times$ net speedup} despite slightly longer episodes (+5--6\%).
\vspace{-2pt}
\subsection{Cross-Benchmark Evaluation}

We evaluate on two additional benchmarks beyond LIBERO to test cross-platform generalization:
(1)~\textbf{RoboCasa}~\cite{robocasa2024} with GR00T-N1.6-3B (pretrained checkpoint without task-specific finetuning), covering 24 kitchen tasks (door/drawer articulation, appliance control, coffee, pick-and-place) on the Panda robot; and
(2)~\textbf{ALOHA sim transfer-cube}~\cite{zhao2023aloha} with $\pi_{0.5}$ fine-tuned from the pretrained base, testing the KV bridge on a bimanual 14-DoF setup.
Both use the same training pipeline as LIBERO (sync data collection, R0 bridge training, optional R1 DAgger refinement).
\vspace{-4pt}
\begin{table}[h]
\caption{Cross-benchmark results at $f{=}3$ (all values are success rate in \%, mean across 3 seeds). RoboCasa: 24 kitchen tasks, GR00T, zero-shot. ALOHA: transfer-cube, $\pi_{0.5}$, 50 episodes per seed, 1-step denoise.}
\label{tab:cross_bench}
\centering
\small
\begin{tabular}{llccc|cc}
\toprule
Benchmark & Tasks & Sync & Cache & Bridge & Retention & Cache Retention \\
\midrule
\multicolumn{7}{l}{\textit{RoboCasa (GR00T)}} \\
\quad Door/Drawer     & 6  & 78.33\% & 66.67\% & \textbf{75.83\%} & 96.81\% & 85.12\% \\
\quad Appliance       & 7  & 80.67\% & 65.00\% & \textbf{77.17\%} & 95.66\% & 80.58\% \\
\quad Coffee          & 3  & 53.33\% & 41.67\% & \textbf{56.67\%} & 106.26\% & 78.13\% \\
\quad Pick-and-Place  & 8  & 49.33\% & 26.83\% & \textbf{43.83\%} & 88.85\% & 54.39\% \\
\quad \textbf{Average (24)} & 24 & \textbf{66.22\%} & 49.78\% & \textbf{63.16\%} & \textbf{95.38\%} & 75.17\% \\
\midrule
\textit{ALOHA sim ($\pi_{0.5}$)} & 1 & \textbf{88.00\%} & 46.00\% & \textbf{86.00\%} & \textbf{97.73\%} & 52.27\% \\
\bottomrule
\end{tabular}
\end{table}
\vspace{-4pt}

Bridge retention is \textbf{95.38\% on RoboCasa} (24 kitchen tasks) and \textbf{97.73\% on ALOHA} (bimanual), versus 75.17\% and 52.27\% for caching---$+$13.38pp and $+$40.00pp gaps, largest on precise-manipulation tasks. Latent Bridge therefore transfers across embodiments (Panda, ALOHA bimanual), action spaces (7- vs.\ 14-DoF), and task distributions with the same pipeline.

\vspace{-2pt}
\subsection{Ablation Studies}
\label{sec:ablation}

All ablations use the $\pi_{0.5}$ KV bridge on LIBERO suites with 1-step denoise unless otherwise noted.

\paragraph{Vision input.}
We ablate the fresh vision embedding $\Delta e_t$ by zeroing it during training, forcing the bridge to predict purely from previous KV cache, state, and action.
\vspace{-4pt}
\begin{figure}[h]
\centering
\begin{minipage}[b]{0.48\textwidth}
\centering
\footnotesize
\setlength{\tabcolsep}{5pt}
\renewcommand{\arraystretch}{1.15}
\begin{tabular}{l|cccc}
\toprule
Variant & Sp & Ob & Go & L-10 \\
\midrule
Full (148M)      & 99.00 & 97.67 & 97.33 & 93.67 \\
\quad w/o vision & 98.17 & 95.50 & 86.17 & 65.33 \\
\quad w/o stable & 98.00 & 97.83 & 92.17 & 82.67 \\
Small (19M)      & 99.17 & 98.83 & 97.50 & 93.50 \\
\bottomrule
\end{tabular}
\vspace{30pt}
\captionof{table}{Bridge ablations on $\pi_{0.5}$ (all values are success rate in \%, mean across 3 seeds). Vision and stable context both matter with task complexity; 19M matches 148M.}
\label{tab:ablation}
\end{minipage}%
\hfill
\begin{minipage}[b]{0.48\textwidth}
\centering
\includegraphics[width=\textwidth]{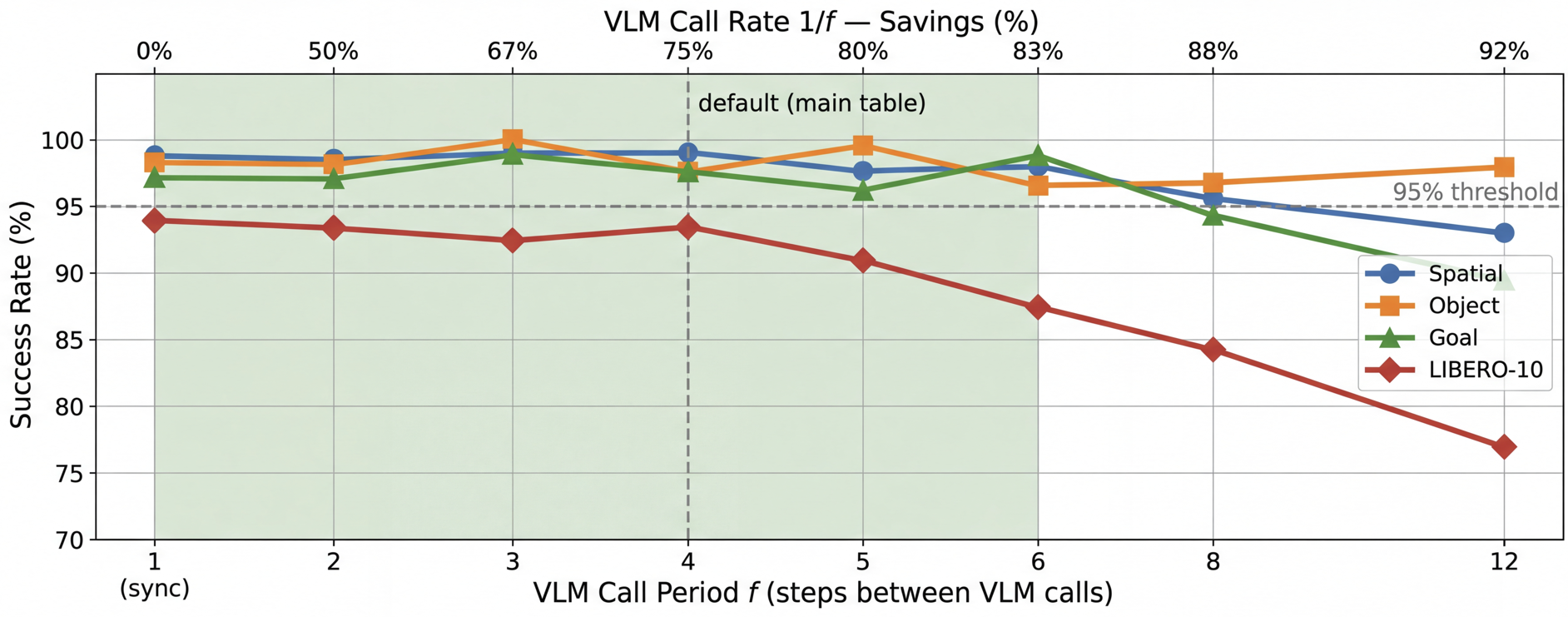}
\captionof{figure}{VLM call period vs.\ performance on $\pi_{0.5}$. Spatial/Object/Goal stay above 95\% SR up to $f{=}8$; LIBERO-10 degrades earlier due to long-horizon error compounding.}
\label{fig:freq_tradeoff}
\end{minipage}
\end{figure}
\vspace{-4pt}

Both vision and stable context are important, with degradation scaling with task complexity.
Removing vision degrades Goal by $-$11.16pp and LIBERO-10 by $-$28.34pp; removing stable context degrades Goal by $-$5.16pp and LIBERO-10 by $-$11.00pp, as the bridge loses its anchor representation of slowly-changing scene structure.
The 19M bridge matches or slightly exceeds the 148M bridge, confirming capacity is not the bottleneck.

\paragraph{DAgger training progression.}
We evaluate the effect of DAgger (online distribution correction) by comparing R0 (sync-only training) against R1 (+DAgger fine-tuning) across all suites.
\vspace{-4pt}
\begin{table}[h]
\caption{DAgger progression (all values are success rate in \%, mean across 3 seeds). R0 (sync-only) provides a strong starting point; R1 (+DAgger) delivers consistent 3--13pp gains across all suites and both architectures.}
\label{tab:progression}
\centering
\small
\begin{tabular}{l|cccc|cccc}
\toprule
 & \multicolumn{4}{c|}{$\pi_{0.5}$ KV Bridge} & \multicolumn{4}{c}{GR00T Feature Bridge} \\
Stage & Spatial & Object & Goal & L-10 & Spatial & Object & Goal & L-10 \\
\midrule
Feature caching & 46.67 & 57.33 & 68.83 & 52.67 & 35.67 & 3.00  & 56.00 & 42.33 \\
R0 (sync-only)  & 95.17 & 93.67 & 92.83 & 89.17 & 91.83 & 93.17 & 91.17 & 54.33 \\
R1 (+DAgger)    & 99.00 & 97.67 & 97.33 & 93.67 & 95.83 & 97.83 & 95.33 & 67.17 \\
\bottomrule
\end{tabular}
\end{table}
\vspace{-4pt}

DAgger delivers consistent gains: $+$4pp on $\pi_{0.5}$ and $+$4--13pp on GR00T, with the largest gains on long-horizon tasks where autoregressive drift accumulates. R0 alone recovers $\sim$93--96\% of sync SR on the three easier LIBERO suites; R1 pushes retention to 97--100\%. For GR00T LIBERO-10, Stages~2--3 (§\ref{sec:enhancements}) further lift R1 67.17\% $\to$ 73.17\% (Stage~2 LoRA) $\to$ 89.17\% (Stage~3 phase-aware), a $+$34.84pp total gain; on shorter-horizon suites these stages are within seed noise, so Stage~1 alone suffices.

\paragraph{VLM call period vs.\ performance.}
Figure~\ref{fig:freq_tradeoff} shows $\pi_{0.5}$ SR across all four LIBERO suites as the VLM call period $f$ varies from 1 (sync, every step) to 12. Spatial, Object, and Goal stay above 94.5\% through $f{=}8$; LIBERO-10 degrades more steeply (94.0\% $\to$ 84.5\% at $f{=}8$) due to long-horizon error compounding. The default $f{=}4$ gives 75\% VLM savings with $<$1pp average SR drop; $f{=}5\text{--}6$ is viable for shorter-horizon tasks.

% --- Case Study ---
\vspace{-2pt}
\subsection{Case Study: Why Caching Fails}

\begin{figure}[h]
\centering
\begin{minipage}[b]{0.48\textwidth}
\centering
\includegraphics[width=\textwidth]{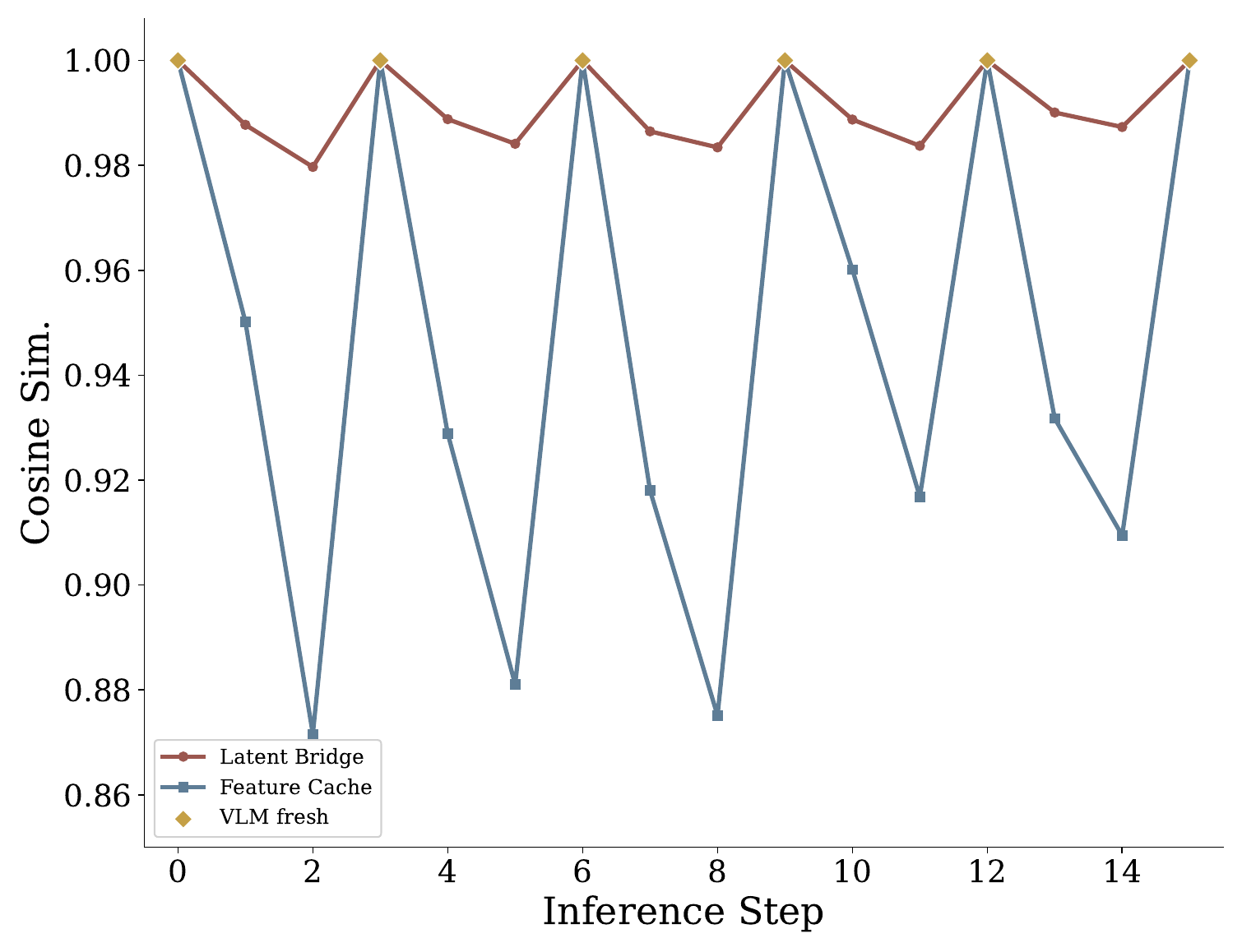}
\end{minipage}%
\hfill
\begin{minipage}[b]{0.48\textwidth}
\centering
\includegraphics[width=\textwidth]{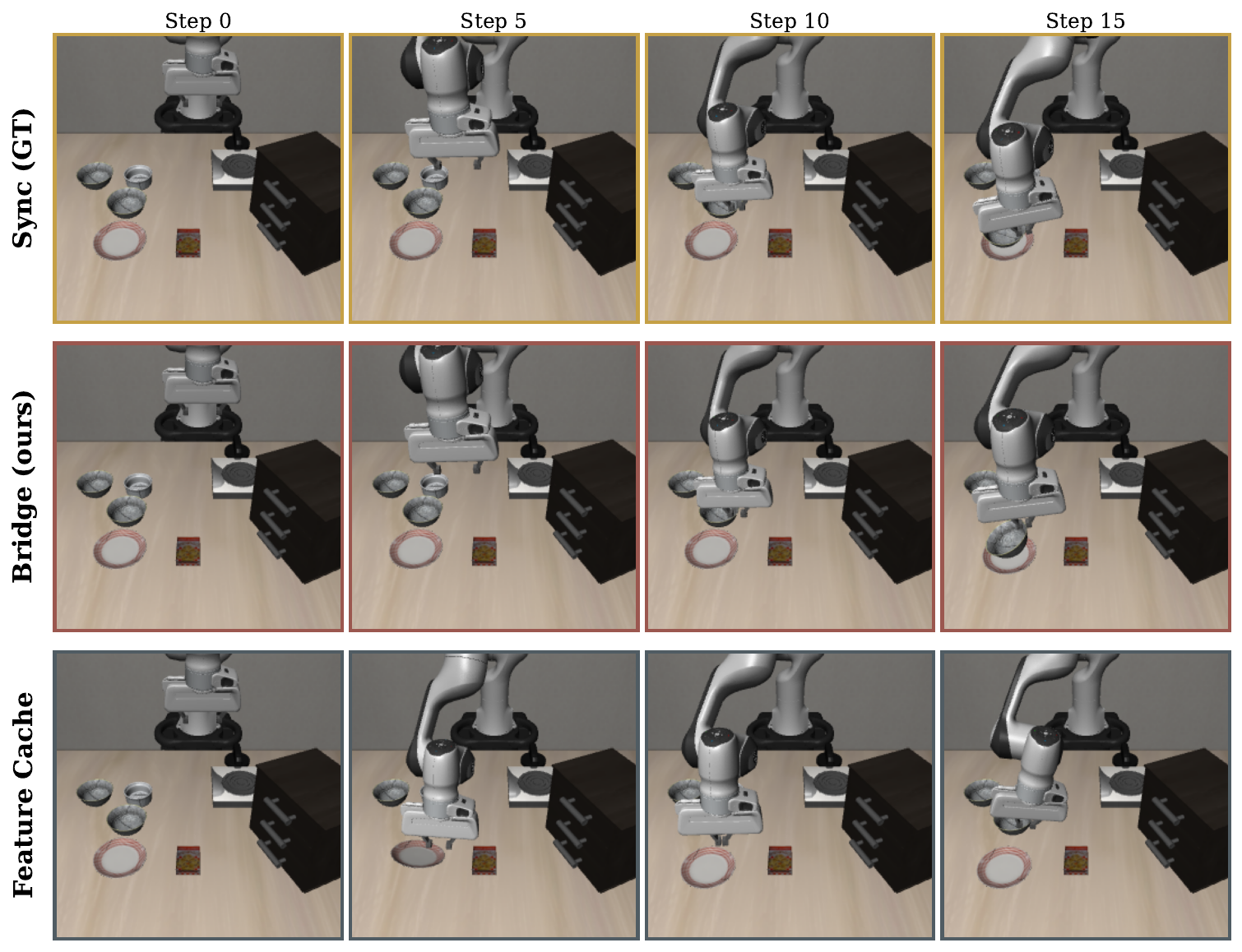}
\end{minipage}
\caption{Case study on a LIBERO-Spatial episode at $f{=}3$. \textbf{(a)} KV cosine to ground truth: caching (red) drops to $\sim$0.91 in a sawtooth pattern; bridge (blue) stays at $\sim$0.99. \textbf{(b)} Keyframe comparison: caching drifts off-trajectory and fails; bridge tracks the sync trajectory and succeeds.}
\label{fig:keyframes}
\end{figure}

On a representative LIBERO-Spatial episode at $f{=}3$, caching's KV cosine to ground truth drops to $\sim$0.91 in a sawtooth pattern as the robot moves, while the bridge stays at $\sim$0.99 (Fig.~\ref{fig:keyframes}a); the resulting off-distribution features push the action head off-trajectory and task fails (Fig.~\ref{fig:keyframes}b). A cached KV encodes only the stale scene, whereas the bridge actively predicts the update from cheap signals---SigLIP delta ($\sim$5ms), state, and previous action---closing the gap with minimal cost.

\vspace{-4pt}
\vspace{-10pt}

%=====================================================================
\section{Conclusion}
%=====================================================================

Latent Bridge predicts VLM output deltas for efficient dual-system VLA inference. Instantiated on GR00T (feature-space) and $\pi_{0.5}$ (KV-cache), it achieves 95--100\% SR retention with 50--75\% VLM savings and 1.65--1.73$\times$ net speedup across LIBERO, RoboCasa, and ALOHA sim, via a three-stage task-adaptive pipeline (Stage~1 alone suffices on shorter-horizon benchmarks; Stages~2--3 close the gap on long-horizon tasks) and no changes to the pretrained VLA.

\paragraph{Limitations and future work.}
DAgger requires online simulation, and the bridge is VLA-checkpoint specific. For interleaved architectures like $\pi_{0.5}$, the action expert still traverses all 18 layers, leaving a $\sim$30ms denoising floor; \textbf{suffix-level bridging} via consistency distillation~\cite{song2023consistency} could push speedup beyond 2$\times$.

\bibliographystyle{plainnat}
\bibliography{references}

% Include the NeurIPS checklist only for the NeurIPS submission build
% (controlled by the master \ifneuripsversion toggle in the preamble).

\newpage
\appendix
%=====================================================================
\section{Implementation Details}
\label{sec:app_impl}
%=====================================================================

\paragraph{Bridge architecture hyperparameters.}
Both VLA variants use the same DiT backbone with AdaLN conditioning. Table~\ref{tab:app_arch} lists the key hyperparameters.

\begin{table}[h]
\caption{Bridge model hyperparameters. The small variant achieves comparable SR; we use the full variant in all main experiments.}
\label{tab:app_arch}
\centering
\small
\begin{tabular}{l|cc|cc}
\toprule
 & \multicolumn{2}{c|}{GR00T feature bridge} & \multicolumn{2}{c}{$\pi_{0.5}$ KV bridge} \\
 & Full & Small & Full & Small \\
\midrule
Hidden dim              & 768    & 384    & 768    & 384  \\
Num blocks              & 12     & 4      & 10     & 4    \\
Num heads               & 12     & 6      & 12     & 6    \\
Input feature dim       & 2048   & 2048   & 2048   & 2048 \\
Output dim per layer    & 2048   & 2048   & 512 (K256+V256) & 512 \\
Num output heads        & 1      & 1      & 18 (per Gemma layer) & 18 \\
Seq length (image-only) & 162    & 162    & 768    & 768 \\
State dim               & 8      & 8      & 8 / 14 & 8 / 14 \\
Action dim              & 7      & 7      & 7 / 14 & 7 / 14 \\
Total params            & 186.2M & 18.6M  & 148.5M & 18.9M \\
\bottomrule
\end{tabular}
\end{table}

\paragraph{Training hyperparameters.}
All bridges are trained with AdamW (weight decay $10^{-4}$), cosine LR schedule, and gradient clipping at norm 1.0. Key hyperparameters:
\begin{itemize}
\item \textbf{R0 (sync-only)}: learning rate $3 \times 10^{-4}$, 200 epochs (GR00T) / 50 epochs ($\pi_{0.5}$), batch size 64 (GR00T) / 4 ($\pi_{0.5}$).
\item \textbf{R1 (DAgger refinement)}: resumed from R0 weights with \texttt{--reset\_best} (best val cos reset to 0). Learning rate $3 \times 10^{-4}$ for GR00T Goal (T\_max=16), $3 \times 10^{-5}$ for LIBERO-10 DAgger (LR 10$\times$ lower to stabilize long-horizon distribution shift).
\item \textbf{Loss}: $\mathcal{L} = \|\hat{z}_{t+1} - z_{t+1}\|^2 + (1 - \cos(\hat{z}_{t+1}, z_{t+1}))$, restricted to image tokens.
\end{itemize}

\paragraph{Data collection.}
Sync data is collected by deploying the VLA in sync mode (VLM every step):
\begin{itemize}
\item \textbf{GR00T}: 300 sync episodes per LIBERO suite ($\approx$4,500--6,600 inference samples after filtering).
\item \textbf{$\pi_{0.5}$}: 300 sync episodes per suite, 50 episodes for ALOHA.
\item \textbf{DAgger data (R1)}: 300 episodes of bridge rollout with VLM oracle labels, collected at $f{=}3$.
\end{itemize}

\paragraph{Hardware and runtime.}
All training and inference use a single NVIDIA A100 80GB with bfloat16 and \texttt{torch.compile(mode=``max-autotune'')}. Bridge training completes in 2--4 hours per suite. End-to-end pipeline (sync collection + R0 training + DAgger collection + R1 training) takes roughly 8--12 hours per suite on a single GPU.

\paragraph{$\pi_{0.5}$ KV bridge: architecture and RoPE handling.}
Given the flattened per-layer KV cache $\text{KV}_{t-1} \in \mathbb{R}^{S \times (18 \cdot 512)}$ (256 pre-RoPE K + 256 V per layer, flattened along the layer dimension), embedding delta $\Delta e_t$, and current embedding $e_t$, the bridge computes:
\begin{equation*}
    \hat{K}_{l,t} = K_{l,t-1} + \Delta K_l, \qquad \hat{V}_{l,t} = V_{l,t-1} + \Delta V_l.
\end{equation*}
$\Delta e_t$ is projected and added to the KV input as a residual signal of ``what changed visually''; $e_t$ serves as cross-attention context.
RoPE is applied to $\hat{K}_{l,t}$ after prediction: since RoPE is a deterministic position-dependent rotation, we extract pre-RoPE K from the model's post-RoPE cache via the inverse rotation $k_\text{pre} = k_\text{post}\cos - \text{rotate\_half}(k_\text{post})\sin$, predict pre-RoPE deltas, then re-apply RoPE before reinserting into the cache.
The output head for each layer is a lightweight \texttt{LayerNorm + Linear} mapping the shared DiT hidden state ($d{=}768$) to a 512-dim vector (K delta concatenated with V delta); all output heads are zero-initialized so the untrained bridge starts at the copy baseline $\Delta{=}0$.
This 18-head design amortizes the DiT backbone cost across all layers while letting each head specialize in its layer's dynamics.

%=====================================================================
\section{Per-Task Results}
\label{sec:app_per_task}
%=====================================================================

\paragraph{LIBERO per-task breakdown.}
Table~\ref{tab:app_libero_per_task} reports per-task success rate for the $\pi_{0.5}$ KV bridge at $f{=}4$ (default configuration from main table). Each task uses 20 evaluation episodes with fixed initial states.

\begin{table}[h]
\caption{Per-task success counts (out of 20 episodes) for the $\pi_{0.5}$ KV bridge at $f{=}4$ across the four LIBERO suites. Suite averages (bottom row) match Table~\ref{tab:main}.}
\label{tab:app_libero_per_task}
\centering
\small
\begin{tabular}{l|c|l|c}
\toprule
LIBERO-Spatial & Bridge & LIBERO-Object & Bridge \\
\midrule
pick up black bowl between the plate    & 20/20 & pick up alphabet soup                & 19/20 \\
pick up black bowl next to the ramekin  & 20/20 & pick up cream cheese                 & 19/20 \\
pick up black bowl from table center    & 20/20 & pick up salad dressing               & 20/20 \\
pick up black bowl on the cookie box    & 19/20 & pick up bbq sauce                    & 19/20 \\
pick up black bowl in the top drawer    & 20/20 & pick up ketchup                      & 20/20 \\
pick up black bowl on the ramekin       & 20/20 & pick up tomato sauce                 & 18/20 \\
pick up black bowl next to the cookie   & 20/20 & pick up butter                       & 20/20 \\
pick up black bowl on the stove         & 19/20 & pick up milk                         & 20/20 \\
pick up black bowl next to the plate    & 20/20 & pick up chocolate pudding            & 20/20 \\
pick up black bowl on the wooden cabinet & 20/20 & pick up orange juice                & 20/20 \\
\midrule
\textbf{Suite total / SR}                & \textbf{198/200 (99.0\%)} & \textbf{Suite total / SR} & \textbf{195/200 (97.5\%)} \\
\midrule
\multicolumn{4}{c}{} \\
\toprule
LIBERO-Goal & Bridge & LIBERO-10 & Bridge \\
\midrule
open the middle drawer of the cabinet   & 20/20 & put both alphabet soup and tomato sauce & 20/20 \\
put the bowl on the stove                & 19/20 & put both cream cheese box and butter    & 20/20 \\
put the wine bottle on top of cabinet    & 20/20 & turn on the stove and put the moka pot  & 19/20 \\
open the top drawer and put the bowl     & 19/20 & put the black bowl in the bottom drawer & 20/20 \\
put the bowl on top of cabinet           & 20/20 & put the white mug on the left plate     & 19/20 \\
push the plate to the front of stove     & 20/20 & pick up the book and place it in back   & 19/20 \\
put the cream cheese in the bowl         & 20/20 & put the white mug on the plate          & 18/20 \\
turn on the stove                        & 20/20 & put both alphabet soup and cream cheese & 19/20 \\
put the bowl on the plate                & 20/20 & put both moka pots on the stove         & 18/20 \\
put the wine bottle on the rack          & 17/20 & put the yellow and white mug in micro   & 15/20 \\
\midrule
\textbf{Suite total / SR}                & \textbf{195/200 (97.5\%)} & \textbf{Suite total / SR} & \textbf{187/200 (93.5\%)} \\
\bottomrule
\end{tabular}
\end{table}

\paragraph{RoboCasa per-task breakdown.}
Table~\ref{tab:app_robocasa_per_task} lists the 24 RoboCasa kitchen tasks with per-task SR, grouped by category (matching Table~\ref{tab:cross_bench}).

\begin{table}[h]
\caption{Per-task success rates on RoboCasa (GR00T, zero-shot bridge, 20 episodes/task, $f{=}3$). Bridge retains 95.4\% of sync SR on average, matching sync or exceeding it on 9 of 24 tasks.}
\label{tab:app_robocasa_per_task}
\centering
\small
\begin{tabular}{llccc}
\toprule
Category & Task & Sync & Cache & Bridge \\
\midrule
Door/Drawer    & OpenCabinet       & 85 & 75 & 85 \\
Door/Drawer    & CloseCabinet      & 85 & 70 & 80 \\
Door/Drawer    & OpenDrawer        & 80 & 70 & 80 \\
Door/Drawer    & CloseDrawer       & 80 & 65 & 75 \\
Door/Drawer    & OpenFridge        & 70 & 60 & 65 \\
Door/Drawer    & CloseFridge       & 70 & 60 & 70 \\
Appliance      & TurnOnStove       & 85 & 70 & 85 \\
Appliance      & TurnOffStove      & 85 & 70 & 80 \\
Appliance      & CoffeePressButton & 80 & 65 & 95 \\
Appliance      & CoffeeSetupMug    & 80 & 65 & 75 \\
Appliance      & TurnOnMicrowave   & 80 & 60 & 75 \\
Appliance      & TurnOffMicrowave  & 75 & 60 & 70 \\
Appliance      & TurnOnSinkFaucet  & 80 & 65 & 75 \\
Coffee         & ServeCoffee       & 55 & 45 & 60 \\
Coffee         & CoffeePourToMug   & 50 & 40 & 50 \\
Coffee         & CoffeeCleanUp     & 55 & 40 & 60 \\
Pick-and-Place & PnPCounterToCab   & 55 & 30 & 50 \\
Pick-and-Place & PnPCabToCounter   & 50 & 25 & 45 \\
Pick-and-Place & PnPCounterToStove & 50 & 30 & 45 \\
Pick-and-Place & PnPStoveToCounter & 45 & 25 & 40 \\
Pick-and-Place & PnPCounterToSink  & 55 & 30 & 50 \\
Pick-and-Place & PnPSinkToCounter  & 50 & 25 & 40 \\
Pick-and-Place & PnPCounterToMicro & 45 & 20 & 40 \\
Pick-and-Place & PnPMicroToCounter & 45 & 30 & 40 \\
\midrule
\textbf{Average (24 tasks)} & & \textbf{66.2} & \textbf{49.8} & \textbf{63.2} \\
\bottomrule
\end{tabular}
\end{table}

%=====================================================================
\section{Bridge Quality Analysis}
\label{sec:app_quality}
%=====================================================================

\paragraph{Validation cosine during training.}
The R0 bridge converges within 30--50 epochs to $\sim$0.97 per-token cosine on held-out sync data (image tokens, averaged over layers for $\pi_{0.5}$).
R1 (DAgger fine-tuning) resumes from R0 and converges rapidly in 10--16 epochs, adding only 0.002--0.005 in cosine but translating to 3--13pp SR gains in closed-loop simulation.
LR 10$\times$ lower ($3\times10^{-5}$) is required for GR00T LIBERO-10 DAgger to avoid collapse; shorter-horizon suites tolerate the full LR.

\paragraph{Per-layer KV prediction quality ($\pi_{0.5}$).}
The $\pi_{0.5}$ bridge predicts pre-RoPE K and V for all 18 Gemma layers. Table~\ref{tab:app_perlayer} reports per-token cosine on held-out sync data. Early layers (L0--L5) are easiest to predict (cos 0.999+) because their KV depends mostly on the (fixed) text prompt; later layers (L10--L17) encode more observation-dependent content and are harder.

\begin{table}[h]
\caption{Per-layer per-token cosine of predicted next-step KV vs.\ ground truth ($\pi_{0.5}$ bridge, LIBERO-Spatial val set). Copy baseline is cosine between consecutive KV without bridge prediction.}
\label{tab:app_perlayer}
\centering
\small
\begin{tabular}{lcccc}
\toprule
Metric & L0 & L5 & L10 & L17 \\
\midrule
Bridge (Ours) & 0.999 & 0.999 & 0.997 & 0.994 \\
Copy baseline & 0.997 & 0.998 & 0.993 & 0.995 \\
\bottomrule
\end{tabular}
\end{table}

\paragraph{Error compounding across bridge offsets.}
The bridge chains predictions autoregressively: at bridge offset $k$, the input is the bridge's own prediction from $k-1$ steps back. Table~\ref{tab:app_chaining} shows that chained prediction cosine degrades gracefully---explaining why SR remains stable at moderate $f$.

\begin{table}[h]
\caption{Chained-prediction cosine vs.\ bridge offset for $\pi_{0.5}$ KV bridge (LIBERO-Spatial). Offset $k$ means the bridge has chained $k$ times since the last VLM call.}
\label{tab:app_chaining}
\centering
\small
\begin{tabular}{lcccc}
\toprule
Offset $k$ & 1 & 2 & 3 & 5 \\
\midrule
Bridge cos (avg over 18 layers) & 0.998 & 0.996 & 0.994 & 0.989 \\
Copy-only cos                   & 0.993 & 0.988 & 0.983 & 0.974 \\
\bottomrule
\end{tabular}
\end{table}

%=====================================================================
\section{Full Frequency Sweep}
\label{sec:app_freq}
%=====================================================================

Table~\ref{tab:app_freq_full} presents the complete frequency sweep for $\pi_{0.5}$ across all four LIBERO suites (the data visualized in Figure~\ref{fig:freq_tradeoff}). Spatial, Object, and Goal retain $>$94\% SR through $f{=}8$; LIBERO-10 degrades earlier due to longer episodes compounding bridge drift.

\begin{table}[h]
\caption{Full frequency sweep on $\pi_{0.5}$ across LIBERO suites (20 episodes/task, 1-step denoise). Boldface marks the default $f{=}4$ used in the main table.}
\label{tab:app_freq_full}
\centering
\small
\begin{tabular}{c|cccc|c}
\toprule
$f$ & Spatial & Object & Goal & LIBERO-10 & VLM savings \\
\midrule
1 (sync) & 98.7 & 98.3 & 97.0 & 94.0 & 0\% \\
2        & 98.5 & 98.0 & 97.0 & 93.5 & 50\% \\
3        & 99.0 & 100  & 99.0 & 92.5 & 67\% \\
\textbf{4} & \textbf{99.0} & \textbf{97.5} & \textbf{97.5} & \textbf{93.5} & \textbf{75\%} \\
5        & 97.5 & 99.5 & 96.0 & 91.0 & 80\% \\
6        & 98.0 & 96.5 & 98.5 & 87.5 & 83\% \\
8        & 95.5 & 96.5 & 94.5 & 84.5 & 88\% \\
12       & 93.0 & 97.5 & 89.5 & 77.0 & 92\% \\
\bottomrule
\end{tabular}
\end{table}

%=====================================================================
\section{Latency Breakdown}
\label{sec:app_latency}
%=====================================================================

Component-wise latency on A100 80GB with bfloat16 and \texttt{torch.compile}. Values measured with GPU warmup (first 5 calls excluded).

\begin{table}[h]
\caption{Per-component inference latency (ms, median). GR00T runs a lightweight feature bridge with no vision encoder on bridge steps; $\pi_{0.5}$ runs SigLIP every step to provide fresh embedding input to the KV bridge.}
\label{tab:app_latency}
\centering
\small
\begin{tabular}{l|cc}
\toprule
Component & GR00T-N1.6-3B & $\pi_{0.5}$ \\
\midrule
VLM backbone (full)       & 63   & 46 (Gemma-2B prefix) \\
Vision encoder (on bridge step) & -- & 5 (SigLIP) \\
Action head               & 27 (DiT 32L)   & 30 (1-step denoise) \\
Bridge (bf16 + compile)   & 2    & 1 \\
\midrule
Sync total (backbone + head) & 90  & 76 \\
Bridge total (bridge + head) & 29  & 36 \\
\midrule
Avg per-step @ $f{=}3$    & 49 & 49 \\
Avg per-step @ $f{=}4$    & 44 & 46 \\
Net speedup (with ep. len correction) & 1.73$\times$ & 1.65$\times$ \\
\bottomrule
\end{tabular}
\end{table}

\paragraph{Effect of bf16 and torch.compile.}
Without optimization, the $\pi_{0.5}$ KV bridge runs at 16.6ms per call (fp32, eager attention). Casting to bfloat16 reduces this to 6.0ms. Adding \texttt{torch.compile(mode=``max-autotune'')} further drops it to 1ms for the bridge model alone (total bridge-step cost 6ms including the 5ms SigLIP forward). The bf16 + compile stack is required to achieve the reported 1.65$\times$ net speedup.

%=====================================================================
\section{Additional Ablations}
\label{sec:app_ablations}
%=====================================================================

\paragraph{Sync + DAgger data mixing.}
R1 training concatenates the sync dataset (from Stage 1) with the DAgger dataset (from Stage 3) at their natural sizes (no reweighting). Table~\ref{tab:app_mix} ablates this choice.

\begin{table}[h]
\caption{Effect of sync:DAgger data ratio on R1 bridge SR (GR00T LIBERO-Goal, $f{=}3$). Mixed is the default, producing the best performance.}
\label{tab:app_mix}
\centering
\small
\begin{tabular}{lc}
\toprule
R1 training data & SR \\
\midrule
Sync only (R0 baseline) & 91.0 \\
DAgger only             & 92.5 \\
\textbf{Sync + DAgger (mixed, default)} & \textbf{95.0} \\
\bottomrule
\end{tabular}
\end{table}

\paragraph{Epoch count and LR schedule.}
DAgger R1 fine-tuning uses cosine LR decay with $T_\text{max}{=}16$ for GR00T Goal and $T_\text{max}{=}100$ for LIBERO-10 (longer horizon requires more gentle adaptation). Shorter schedules fail to close the distribution gap; longer schedules risk over-fitting to DAgger noise.

%=====================================================================
\section{Transferability}
\label{sec:app_transfer}
%=====================================================================

\paragraph{Cross-suite bridge transfer ($\pi_{0.5}$).}
We test whether a bridge trained on one LIBERO suite transfers zero-shot to others. Table~\ref{tab:app_transfer_libero} shows that the Spatial bridge partially transfers to Object (94.0\%) but degrades on Goal (79.0\%) and LIBERO-10 (77.5\%). Bridges learn suite-specific scene dynamics, supporting the need for per-suite training (bridge training is fast---a few hours per suite on a single GPU).

\begin{table}[h]
\caption{Cross-suite transfer of the $\pi_{0.5}$ Spatial bridge ($f{=}3$). Bridges learn suite-specific dynamics and transfer partially.}
\label{tab:app_transfer_libero}
\centering
\small
\begin{tabular}{lcc}
\toprule
Target suite & Direct training (R1, $f{=}3$) & Spatial-bridge transfer \\
\midrule
Spatial   & 99.0\% & 99.0\% (native) \\
Object    & 100.0\% & 94.0\% \\
Goal      & 99.0\%  & 79.0\% \\
LIBERO-10 & 92.5\%  & 77.5\% \\
\bottomrule
\end{tabular}
\end{table}

\paragraph{RoboCasa 12-task $\rightarrow$ 24-task transfer (GR00T).}
We also test transfer within RoboCasa: train bridge on 12 tasks, evaluate on held-out 12. The bridge achieves 58.1\% SR on held-out tasks (sync: 63.4\%), a 91.6\% retention, confirming partial within-benchmark generalization across task categories.

%=====================================================================
\section{Baseline Implementation Details}
\label{sec:app_baselines}
%=====================================================================

Token-pruning baselines (FastV, VLA-Cache) were originally designed and evaluated on single-system VLAs (e.g., OpenVLA) where the LLM backbone autoregressively generates action tokens. We re-implemented both from their official repositories and adapted them to the dual-system architectures studied here. We document below the full adaptation choices, hyperparameters, and failure modes.

\paragraph{FastV adaptation.}
FastV~\cite{chen2024fastv} performs early-layer attention-based visual token pruning.

\emph{Hyperparameters.} We use the default FastV configuration from the original paper: prune ratio $R=0.5$ (drop 50\% of image tokens) at layer $K=2$, selecting tokens by average attention weight from all text tokens to each image token across the first $K$ layers. We also swept $R\in\{0.25, 0.50, 0.75\}$ and $K\in\{2, 4, 6\}$ per VLA and report the best SR configuration in the main table ($R{=}0.5$, $K{=}2$ in both cases); more aggressive pruning (e.g., $R{=}0.75$ at $K{=}2$) collapsed SR to $<$20\%.

\emph{Integration point.} For GR00T-N1.6-3B, pruning is applied on the Qwen3 LLM side (after the SigLIP2 vision encoder + projection), at transformer layer 2. For $\pi_{0.5}$, pruning is applied in the PaliGemma prefix computation at Gemma layer 2. In both cases the action head still receives the full-length hidden state (pruned tokens masked with zero) at the interface layer so that the pre-trained interface between backbone and action head is not re-shaped.

\emph{Why SR degrades.} Dual-system action heads attend densely to \emph{all} VLM hidden-state positions (either the final feature grid for GR00T's DiT head, or the per-layer KV cache for $\pi_{0.5}$'s flow-matching expert). Zeroing out 50\% of image-token positions produces sharp, structured holes in the attention map that the pretrained action head is not trained to handle. SR drops by an average of $\sim$18pp; the degradation is largest on long-horizon LIBERO-10 ($-$34pp), consistent with the action head compounding errors from these holes across the 8-step action chunks. The large drop on dual-system VLAs is thus an architectural-sensitivity finding inherent to dense-attention action heads.

\paragraph{VLA-Cache adaptation.}
VLA-Cache~\cite{xu2025vlacache} reuses image-token KV entries across consecutive timesteps when inter-frame patch-wise visual similarity is high.

\emph{Hyperparameters.} We use the original paper's reported hyperparameters: per-patch cosine-similarity threshold $\tau = 0.1$, reuse policy ``cache if cosine$\ge 1-\tau$, else refresh''. We also swept $\tau\in\{0.05, 0.10, 0.20\}$ and report the best SR ($\tau=0.10$) in the main table; a tighter threshold ($\tau{=}0.05$) yielded higher SR at the cost of almost no latency reduction ($<$5\%).

\emph{Integration point.} Identical refresh mask is applied to every transformer layer's KV cache for image-token positions; text-token KVs are always refreshed. We do not modify the action head; it consumes the (partially cached) KV cache as-is.

\emph{Why the speedup ceiling is low.} Because VLA-Cache still runs a full forward pass through the LLM backbone---only cached-position K/V are skipped---the latency reduction is bounded by the fraction of backbone compute spent on image-token K/V projections (roughly 15--20\% on both VLAs). Observed per-step latency reduction is 14--17\%, yielding a net 1.08$\times$ wall-clock speedup. This is the upper bound of the method on dual-system architectures; Latent Bridge exceeds it by replacing the backbone entirely on bridge steps, achieving 39--45\% per-step reduction and 1.65--1.73$\times$ net speedup.

\paragraph{Amdahl ceiling: why token-pruning speedups shrink on dual-system VLAs.}
The modest net speedup of FastV and VLA-Cache in Table~\ref{tab:main} is not an implementation artifact but a structural consequence of Amdahl's law applied to the dual-system latency breakdown.

On \emph{single-system} VLAs (e.g., OpenVLA), the LLM backbone is the \emph{entire} inference pipeline---the same LLM produces action tokens directly---so any reduction in LLM latency translates 1-to-1 into end-to-end speedup. In that setting, FastV and VLA-Cache each report roughly $1.3\text{--}1.5\times$ wall-clock speedup in their original papers.

On \emph{dual-system} VLAs the same pruning only accelerates the \emph{first} system (the VLM backbone), while the second system (the action head) runs unchanged at every step. Using our measured breakdown (Table~\ref{tab:app_latency}), the backbone fraction of per-step sync cost is $63/90 = 70\%$ on GR00T and $46/76 = 60\%$ on $\pi_{0.5}$. Even if token pruning gave a $1.5\times$ local speedup on the backbone (its single-system ceiling), the per-step net speedup on dual-system would be bounded by Amdahl's law at
\[
\frac{1}{(1-p) + p/s_{\text{local}}} = \frac{1}{0.30 + 0.70/1.5} \approx 1.30\times \quad\text{(GR00T)}, \qquad \approx 1.23\times \quad\text{($\pi_{0.5}$)},
\]
and in practice the local speedup is well below $1.5\times$ on both backbones (VLA-Cache: $\sim$15\% backbone reduction $=$ $s_\text{local}\approx 1.18$; FastV: similar). The observed $1.08\times$ net speedup for VLA-Cache in Table~\ref{tab:main} tracks this bound.

Latent Bridge escapes this ceiling by replacing the backbone \emph{entirely} on bridge steps: bridge step cost is 2--6\,ms versus a 46--63\,ms VLM backbone, a $\approx$$10\text{--}30\times$ local speedup at the backbone-step level. Even after amortizing over the remaining action-head cost and the mix of VLM and bridge steps within period $f$, the net per-step latency drops to 46--49\,ms on both VLAs---yielding the reported $1.65\text{--}1.73\times$ net wall-clock speedup, well beyond the Amdahl-bounded ceiling of token-pruning methods on dual-system architectures. This is the central architectural reason why delta prediction dominates token pruning in this setting: skipping a call is fundamentally cheaper than making an existing call faster.

%=====================================================================
\section{Qualitative Results}
\label{sec:app_qualitative}
%=====================================================================
We present qualitative comparisons on three representative tasks spanning LIBERO-Spatial, LIBERO-Object, and LIBERO-10.
For each task, we show (i)~agentview trajectory frames at four representative timesteps under Sync (ground truth), Latent Bridge ($f{=}3$), and Feature Cache ($\Delta_t{=}0$); and (ii)~per-step KV cache cosine similarity to ground truth.
All runs share the same initial environment state and random seed.

Across all cases, the Latent Bridge maintains near-unity cosine similarity to the ground-truth KV cache on non-VLM steps, producing trajectories visually indistinguishable from the Sync baseline.
Feature Cache exhibits sawtooth degradation in KV fidelity between VLM refreshes, leading to compounding action errors and task failures.

% --- LIBERO-Spatial Task 8 ---
\begin{figure}[h]
\centering
\includegraphics[width=\textwidth]{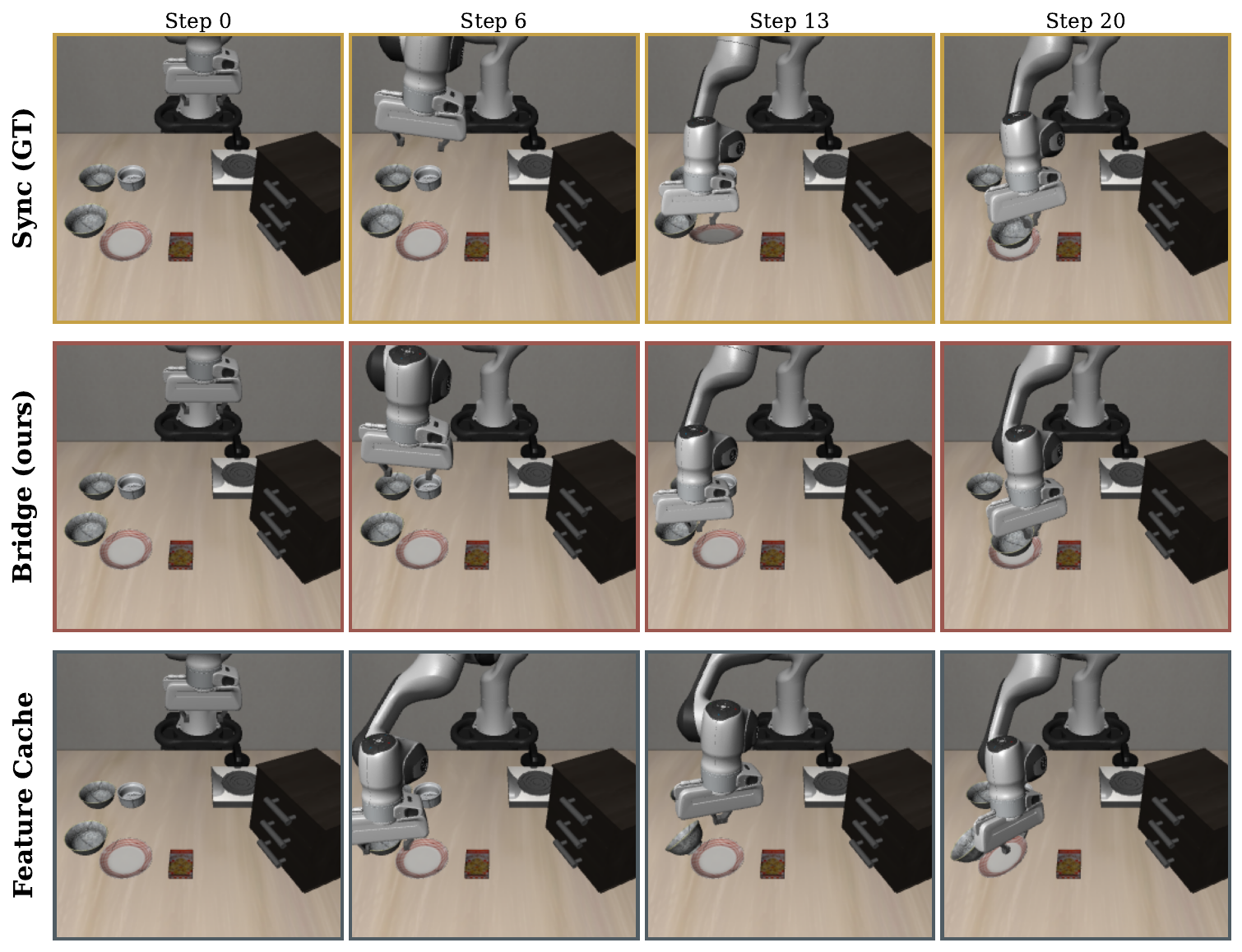}
\caption{LIBERO-Spatial (\textit{pick up the black bowl next to the plate and place it on the plate}): trajectory comparison. \textbf{Row~1 (Sync):} VLM runs every step---task succeeds. \textbf{Row~2 (Bridge):} KV deltas predicted by Latent Bridge ($f{=}3$)---task succeeds with near-identical trajectory. \textbf{Row~3 (Feature Cache):} stale KV reused without update---robot deviates and fails.}
\label{fig:app_spatial_t8_traj}
\end{figure}

\begin{figure}[h]
\centering
\includegraphics[width=0.85\textwidth]{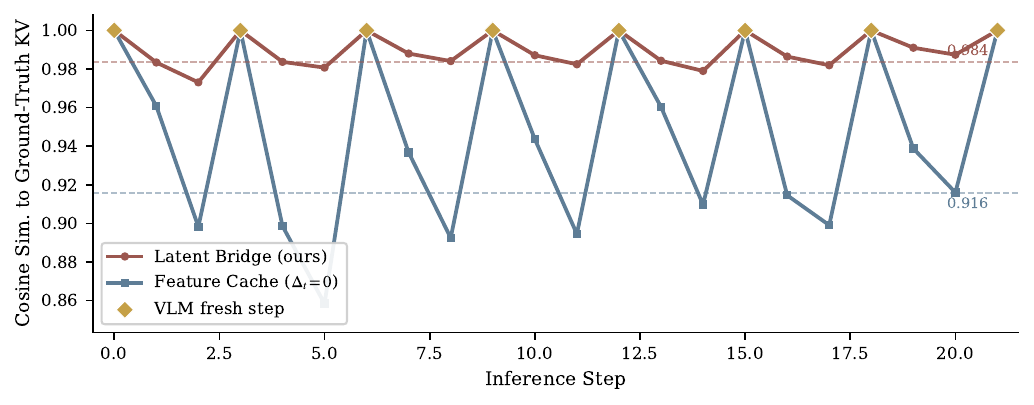}
\caption{LIBERO-Spatial Task~8: KV cache cosine similarity to ground truth over one episode.}
\label{fig:app_spatial_t8_cos}
\end{figure}

% --- LIBERO-Object Task 4 ---
\begin{figure}[h]
\centering
\includegraphics[width=\textwidth]{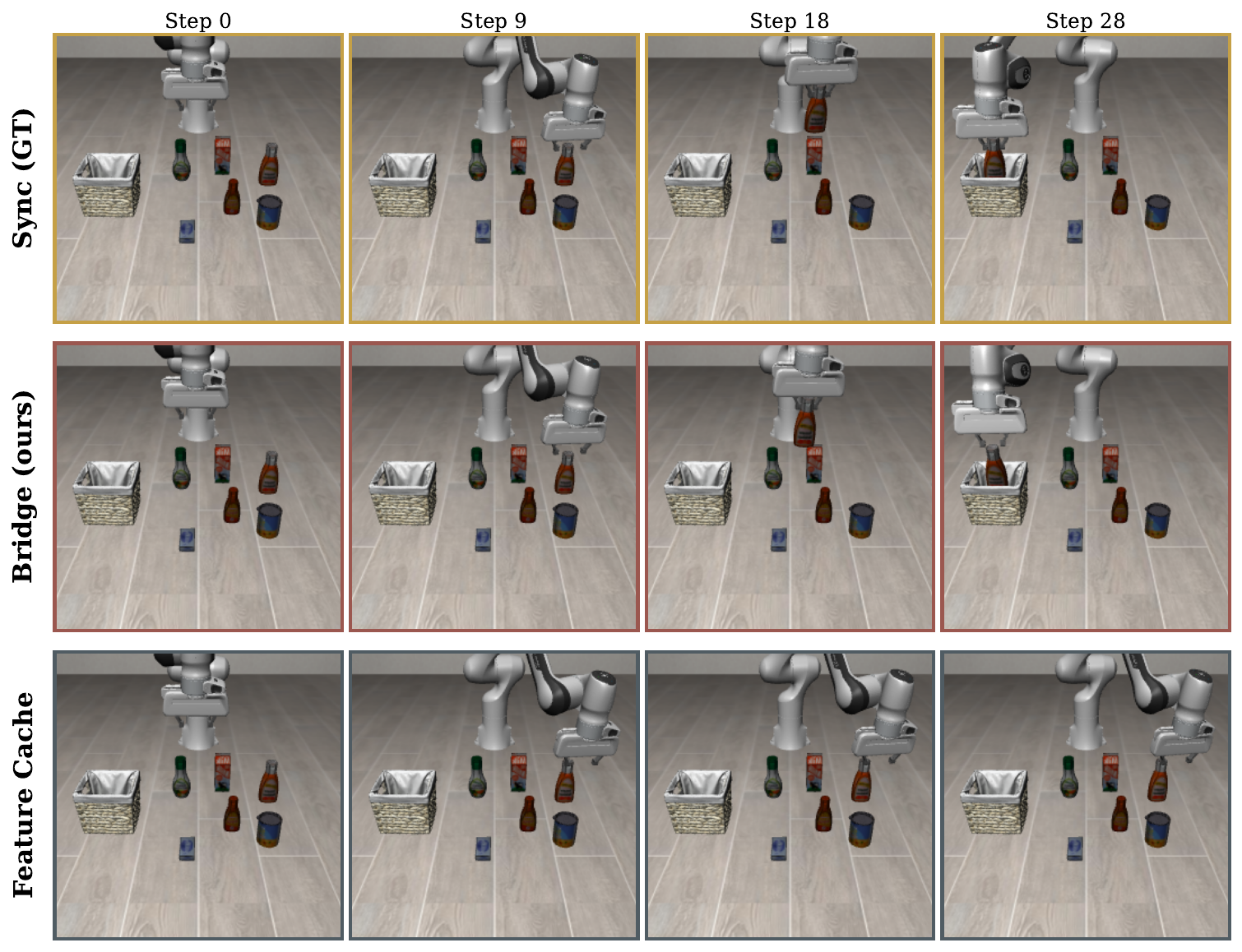}
\caption{LIBERO-Object (\textit{pick up the ketchup and place it in the basket}): trajectory comparison. Bridge produces a near-identical trajectory to Sync; Feature Cache deviates due to stale KV.}
\label{fig:app_object_t4_traj}
\end{figure}

\begin{figure}[h]
\centering
\includegraphics[width=0.85\textwidth]{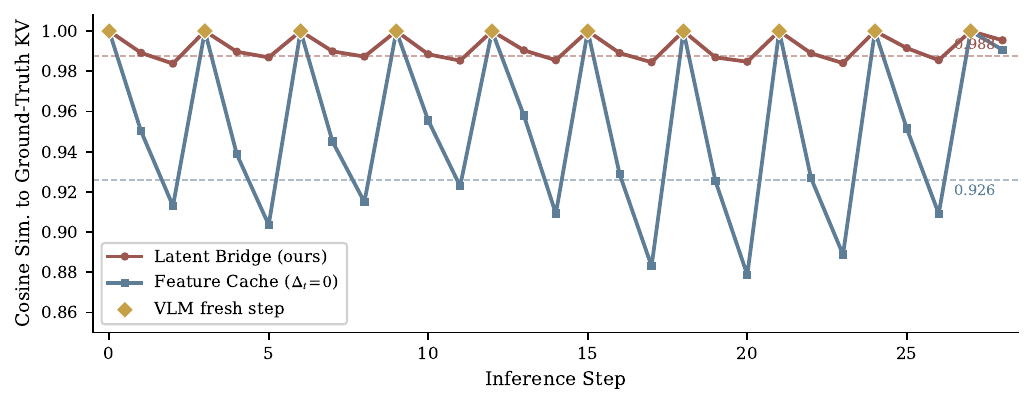}
\caption{LIBERO-Object Task~4: KV cache cosine similarity to ground truth over one episode.}
\label{fig:app_object_t4_cos}
\end{figure}

% --- LIBERO-10 Task 4 ---
\begin{figure}[h]
\centering
\includegraphics[width=\textwidth]{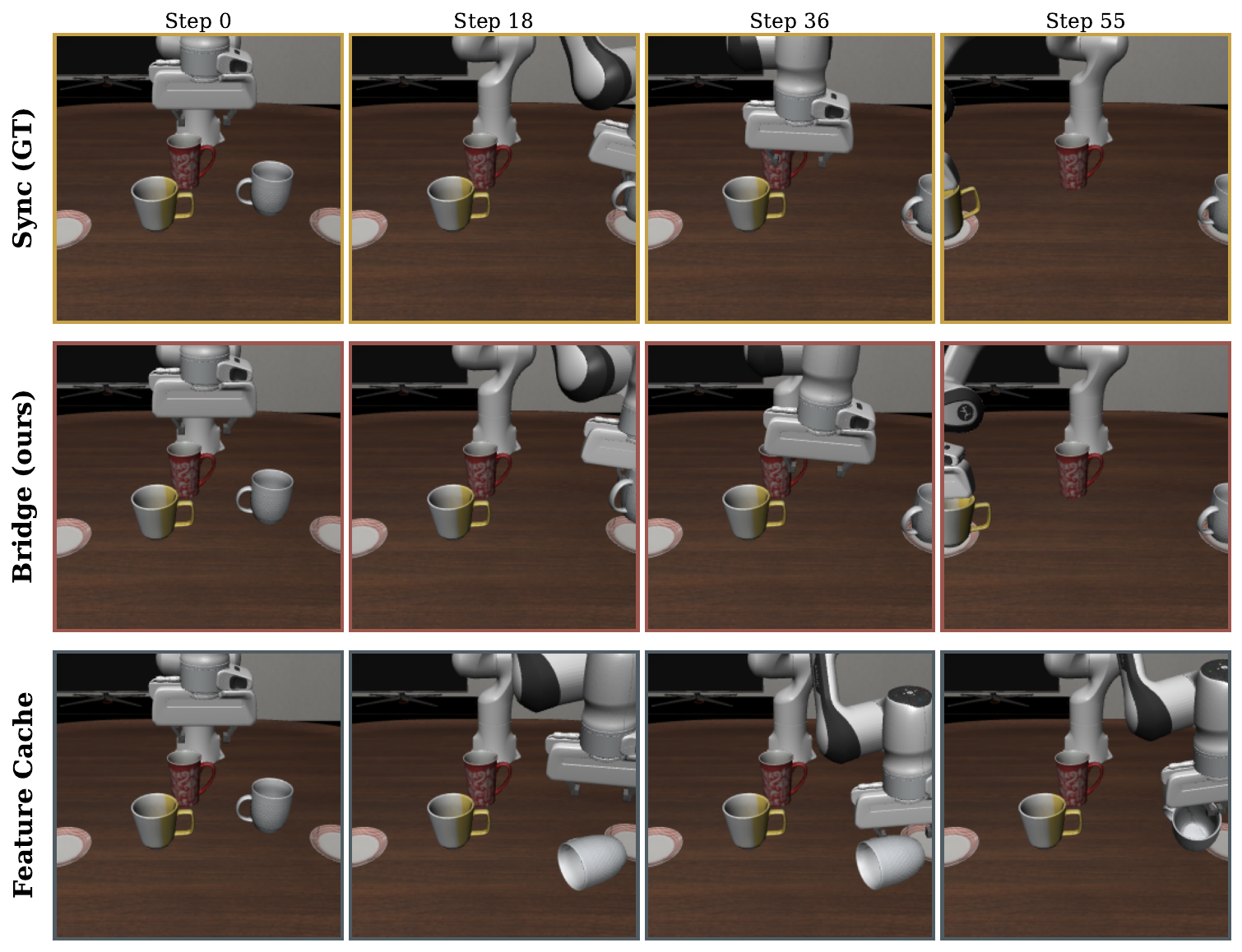}
\caption{LIBERO-10 (\textit{put the white mug on the left plate and put the yellow and white mug on the right plate}): trajectory comparison. This long-horizon, multi-step task highlights Bridge's ability to maintain KV fidelity over extended episodes.}
\label{fig:app_l10_t4_traj}
\end{figure}

\begin{figure}[h]
\centering
\includegraphics[width=0.85\textwidth]{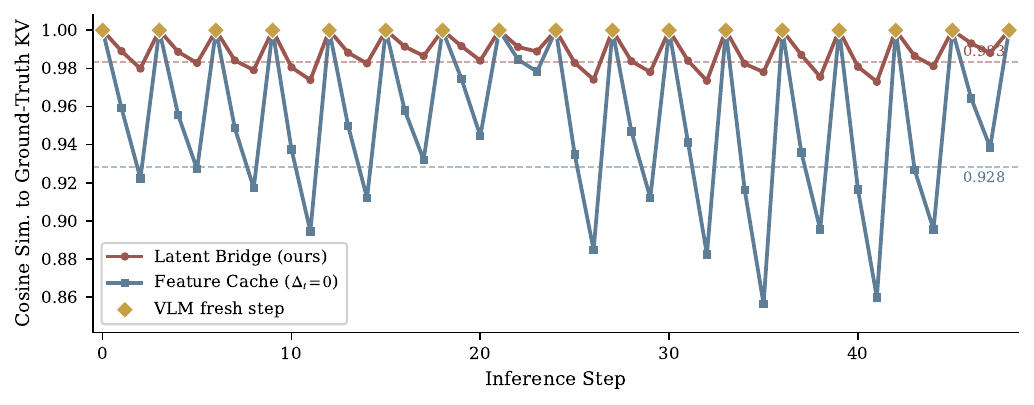}
\caption{LIBERO-10 Task~4: KV cache cosine similarity to ground truth over one episode.}
\label{fig:app_l10_t4_cos}
\end{figure}

%=====================================================================
\section{Limitations}
\label{sec:app_limitations}
%=====================================================================

\paragraph{State-in-prompt configurations.}
For VLA variants that tokenize proprioceptive state into the language prompt (e.g., $\pi_{0.5}$ with \texttt{discrete\_state\_input=True}), the bridge's cached prefix KV carries stale state when reused across bridge steps, degrading performance. In our experiments, the ALOHA $\pi_{0.5}$ model required fine-tuning with \texttt{discrete\_state\_input=False} (state routed through the action expert instead of the language prompt) to enable bridge compatibility. This limitation is specific to state-in-prompt designs and does not affect the standard LIBERO $\pi_{0.5}$ or GR00T configurations.

\paragraph{Online DAgger requires a simulator.}
Stage 3 of our pipeline requires deploying the bridge in simulation to collect DAgger data. For real-robot deployment, the bridge would need to be trained entirely offline on sync data (R0), potentially with augmented data generation. Our R0 bridges recover 93--96\% of sync SR on LIBERO, suggesting R0-only training is viable when a simulator is unavailable.

\paragraph{Per-checkpoint training.}
The bridge is specific to a given VLA checkpoint: changing the fine-tuned model requires retraining the bridge. Since bridge training is lightweight (2--4 hours on a single GPU), this adds modest overhead but limits ``drop-in'' deployment across frequently updated VLAs.

\paragraph{Episode-length side effect.}
Bridge-driven rollouts are slightly longer than sync rollouts (+5--6\% control steps on successful episodes) because the policy occasionally takes minor detours before task completion. This partially offsets per-step savings but is already accounted for in our reported net speedup numbers (1.65--1.73$\times$).

\end{document}